# STOCHASTIC CAPACITATED ARC ROUTING PROBLEM

Gérard Fleury[1]
Lacomme Philippe[2]    Christian Prins[3]

[1] Université Blaise Pascal
Laboratoire d'Informatique (LIMOS) UMR CNRS 6158,
Campus des Cézeaux, 63177 Aubiere Cedex
fleury@math.univ-bpclermont.fr

[2] Université Blaise Pascal
Laboratoire d'Informatique (LIMOS) UMR CNRS 6158,
Campus des Cézeaux, 63177 Aubiere Cedex
placomme@sp.isima.fr

[3] Université de Technologie de Troyes,
ISTIT (équipe OSI) FRE CNRS 2732
12, Rue Marie Curie, BP 2060, F-10010 Troyes Cedex (France)
prins@utt.fr

# STOCHASTIC CAPACITATED ARC ROUTING PROBLEM[*]

**GÉRARD FLEURY**

Blaise Pascal University, Applied Mathematics Laboratory (UMR CNRS 6620), 63177 Aubière (France)

**PHILIPPE LACOMME**

Blaise Pascal University, Computer Science Laboratory (UMR CNRS 6158), 63177 Aubière (France)

**CHRISTIAN PRINS**

University of Troyes, Computer Science Laboratory, 12 rue Marie-Curie, 10010 Troyes (France)

# Table of contents

# Index of figures

# Index of tables

# Index

# STOCHASTIC CAPACITATED ARC ROUTING PROBLEM

**GÉRARD FLEURY**

Blaise Pascal University, Applied Mathematics Laboratory (UMR CNRS 6620),
63177 Aubière (France)

**PHILIPPE LACOMME**

Blaise Pascal University, Computer Science Laboratory (UMR CNRS 6158),
63177 Aubière (France)

**CHRISTIAN PRINS**

University of Troyes, Institut des Sciences et Technologies de l'Information de Troyes,
12 rue Marie-Curie, 10010 Troyes (France)

This paper deals with the Stochastic Capacitated Arc Routing Problem (SCARP), obtained by randomizing quantities on the arcs in the CARP. Optimization problems for the SCARP are characterized by decisions that are made without knowing their full consequences. For real-life problems, it is important to create solutions insensitive to variations of the quantities to collect because of the randomness of these quantities. Efficient robust solutions are required to avoid unprofitable costly moves of vehicles to the depot node. Different criteria are proposed to model the SCARP and advanced concepts of a genetic algorithm optimizing both cost and robustness are provided. The method is benchmarked on the well-known instances proposed by DeArmon, Eglese and Belenguer. The results prove it is possible to obtain robust solutions without any significant enlargement of the solution cost. This allows treating more realistic problems including industrial goals and constraints linked to variations in the quantities to be collected.

The concern of this paper is the *Stochastic Capacitated Arc Routing Problems* (SCARPs) which are extensions of the *Capacitated Arc Routing Problem* (CARP). The CARP is a distribution problem introduced by Golden and Wang (1981) and defined on an undirected network $G = (V, E)$ where $V$ is a set of $n$ nodes and $E$ a set of $m$ edges. A fleet of vehicles of a given capacity $Q$ is based in a depot (a distinguished node) $s$. A subset $R$ of $t$ required edges or *tasks* must be serviced by one vehicle. Any edge can be traversed several times but each edge $(i, j)$ has a positive cost $c_{ij} > 0$ and a nonnegative edge demand $q_{ij} \geq 0$ never exceeding the capacity $Q$ of the vehicles: $0 \leq q_{ij} \leq Q$. The CARP consists in determining vehicle trips of minimal total cost containing the depot node and such that each required edge (i.e. with a positive demand $q_{ij} > 0$) is serviced by exactly one vehicle.

The cost of a trip is the sum of the costs of its serviced edges and of its intermediate connecting paths. The applications of CARP come out in a wide range of practical routing problems including: urban waste collection, snow plowing, sweeping and gritting. The undirected network permits to tackle both sides of roads which can be serviced during one traversal and in any direction. Using a directed network, each arc can model one street (or one side of the street) with a fixed direction for service. Let us note the difference between an edge $(i, j)$ and a pair of opposite arcs $(i, j)$ and $(j, i)$ for street networks: for a pair of arcs, each side needs to be serviced separately. Both the directed and undirected CARP is NP-hard. Furthermore the basic-vehicle routing problem named *Rural Postman Problem* is NP-hard.

In real life problems (urban waste collection for example), the demands can not be exactly identified before the trips are planned. The demands are modeled by random variables, i.e. the positive constants $q_{ij} > 0$ are replaced by positive random variables $Q_{ij} > 0$. The basic deterministic CARP (DCARP to avoid any ambiguity) then becomes a stochastic problem (SCARP). The problem is to find solution $S$ expected to perform well (having a low cost) and proving to be robust when facing some variations of demands due to random events.

To our knowledge, Fleury *et al.* (2002), Fleury *et al.* (2004) Fleury *et al.* (2005) report the only previous published works on the SCARP. A heuristic strategy based on a Hybrid Genetic Algorithm (HGA) is implemented to improve the robustness of solutions. However they propose a sensitivity analysis of solutions to arc routing problems but no optimization search strategy to improve robustness: they only solve a customized DCARP in the resolution process. The purpose of this paper is to present methods to efficiently solve SCARPs, determining high-quality solutions as regards the consequences of demand fluctuations. The use of Hybrid Genetic Algorithms (HGA) has been investigated because the HGA (proposed in Lacomme *et al.* (2001)) has been proved to be a very efficient method for the DCARP providing optimal or nearly optimal solutions with a reasonable amount of computational time.

The rest of this paper is organized as follows. First, SCARP problems are defined and a mathematical study is provided under hypotheses on demand and vehicles management rules. A general framework composed of two phases (an optimization phase and a replication phase) is proposed. The key composed of the framework consists in the optimization phase based on the genetic algorithm first introduced by (Lacomme *et al.*, 2001) and the mathematical formula. Finally, the robustness of the solutions obtained by the proposed approach is evaluated on three sets of benchmarks from literature.

## 1 STOCHASTIC CARP

### 1.1 Stochastic CARP definition

A SCARP problem is similar to a DCARP problem, except that positive demands $q_{ij}$ then become positive random variables $Q_{ij}$. To any SCARP problem, can be associated a DCARP problem, where the stochastic demands $Q_{ij}$ are replaced by their expectation $q_{ij} = \overline{Q_{ij}}$. To avoid any ambiguity we call "stochastic" an element applied to the SCARP and "deterministic" an element applied to its associated DCARP. The objective in the DCARP consists in determining a set of trips (solutions) of minimal cost and the objective solving the SCARP consists in determining a robust set of trips (a robust solution). Based on the definition of Jensen (2001), robust solutions are solutions expected to perform well after variations in quantities to collect. Let us note:

$X$ the (finite) set of all solutions.

$S$ a solution of the CARP (finite set of trips).

$Q$ the common capacity of the vehicles used.

$c_{ij}$ the (deterministic) cost of the edge *(i,j)* ($c_{ij} \geq 0$).

$q_{ij}$ the deterministic demand on edge *(i,j)* ($0 \leq q_{ij} \leq Q$).

$Q_{ij}$ the random demand on edge *(i,j)* ($0 \leq Q_{ij} \leq Q$) and if $q_{ij} > 0$ then $Q_{ij} > 0$.

$t(S)$ the deterministic number of trips of $S$.

$T(S)$ the random number of trips of $S$.

$\overline{T(S)}$ the average number of trips of $S$.

$G_j$ the trip *j* of the solution (finite sequence of connected edges, beginning and ending at *s*).

$h(S)$ the deterministic cost of a solution $S$.

$H(S)$ the random variable being the cost of $S$ (sum of the costs of every edge used by each vehicle, with, sometimes, extra trips to the depot node…).

$\overline{H(S)}$ the average cost of $S$.

Let us consider a (deterministic) solution. If the random demands serviced by a trip become less important than expected, the cost does not vary, but, if the vehicle exhausts its capacity before the end of the trip, it is assumed to move from its current position in the network to the depot node and to return to this position to complete the trip initially planned. Such operations create an extra trip and imply a (possibly huge) increasing of the total cost. Such events occur in many applications like waste collection. Calling another vehicle may be impossible for several reasons: the driver can not inform his colleagues due to the lack of communication systems, the other drivers can not come because all trips are performed in parallel, the driver is the only one knowing this sector, and so on. Therefore, for any solution $S$, one has $H(S) \geq h(S)$.

Let us consider, for example, a DCARP instance with only 9 tasks, 3 vehicles with $Q = 4$ and demands equal to 1 for each task. Assume a solution with a deterministic cost $H(S) = 100$ and 3 trips (figure 1). The total loads of vehicles 1,2,3 are respectively equal to 3 (serviced tasks 7,8,9), 2 (tasks 1,2) and 4 (tasks 3,4,5,6).

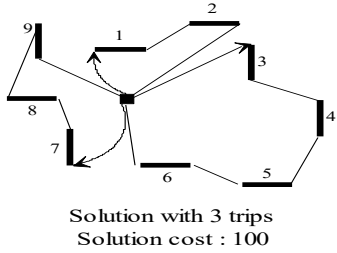

Solution with 3 trips
Solution cost : 100

**Figure 1. A solution for the DCARP**

Random events occurring in practice may affect estimated demands. Assume that after task 4, vehicle load is in fact 3.5 and the actual demands for tasks 5 and 6 become 1.2 and 1.4.

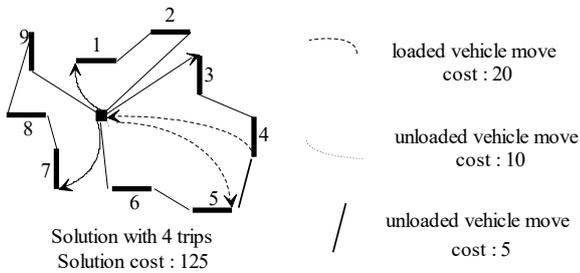

Solution with 4 trips
Solution cost : 125

**Figure 2. The same solution for the SCARP**

Because the vehicle can not service task 5, it moves from its current position (end of task 4) to the depot node and moves back to the beginning node of the task 5 to complete its trip (figure 2). This trip can be viewed as an additional trip. Due to this unproductive move, the solution cost becomes $H(S,\omega) = 125$ when $h(S) = 100$. And one has $T(S,\omega) = 4$ when $t(S)=3$.

### 1.2 Modelization of random quantities in the Stochastic CARP

Random quantities $Q_{ij}$ to be collected on edge $(i, j)$ follow Gaussian laws. In many practical applications, the demand on each edge is the sum of a large number of elementary demands, for instance waste containers or plastic bags in municipal waste collection. But, when each measurement is the result of a large amount of small, independent error sources (having finite mean and variance), the central limit theorem applies and the random demand can be efficiently modeled by Gaussian random variables $N(q_{ij}, \sigma_{ij}^2)$ with two restrictions: a demand must be positive and can not exceed the capacity $Q$ of the vehicles. Hence, $Q_{ij}$ is truncated to avoid the rare values inferior to 1 or superior to $Q$. Such rare events may occur when a lot of trials are done.

### 1.3 Literature on stochastic routing

The literature on stochastic vehicle routing concentrates on the vehicle routing problem on nodes, with random customers and/or demands (Stochastic Vehicle Routing Problem), see for instance Gendreau *et al.* (1996a) for a review, and Laporte and Louveaux (1999), Verweij et al (2001) for additional information.

There are two main approaches for a direct solving of this stochastic problems. In Chance Constrained Programming (CCP), some constraints are satisfied with given probabilities. In Stochastic Programming with Recourse (SPR), a preliminary solution is computed without knowing the realizations of the random variables. Later, a corrective action (recourse) is applied when a constraint is violated. The second class of approaches is based on both single pass heuristics methods and iterative methods (tabu search for example).

A great number of papers deal with the Stochastic VRP with stochastic demands and/or random arc capacities (see for example (Bertsimas *et al.*, (1996), Laporte *et al.*, (2002), Raymond and Powell (1996)). To the best of our knowledge, no paper has been published for the SCARP. Most SCARP applications also require preliminary solutions: for instance, in urban waste collection, the collecting days in the week must be announced in advance to the population. Theoretically, a SCARP could be transformed into a SVRP and solved via SPR. In fact, our HGA for the CARP was designed for large instances (190 for instance) and extensions like forbidden turns and mixed graphs. The conversion into a VRP increases the problem size in such conditions: e.g., each node must be replaced by 4 artificial nodes to handle forbidden turns properly. Since the largest problem solved via SPR has 70 customers, SPR can not be envisaged. One can note some very promising works achieved in problems close to the VRP. See for example Swihart and Papastavrou (1999) for their work on the single-vehicle pick-up and delivery problem, Gendreau *et al.*

(1996b) for a tabu search experimentation for the VRP and Cheung and Powell (1996) for a study of the expected recourse function based on successive convex approximation.

Since problems of interest in arc routing are large scale instances the design of robust metaheuristics is promising. In Fleury *et al.* (2001) for instance, a stochastic metaheuristic is proposed to optimize a random variable without any restrictive assumption on its probability distribution. Robust genetic algorithms have also been studied and applied to the functions of two variables: Tsutsui and Ghosh (1997) add a noise to the objective function, while Branke (Branke, 1998) performs a quick simulation to evaluate the robustness of each child.

To the best of our knowledge, there are only two previous published papers on the Stochastic CARP. First, Fleury *et al.* (2002) propose a robustness evaluation of CARP solutions. Second, Fleury *et al.* (2003, 2004) propose a heuristic based approach (with no iterative searching scheme devoted to the SCARP) to compute robust solutions. The proposed approach is limited because there is no explicit minimization of a criteria linked to the robustness of solutions.

Our goal is to design a metaheuristics devoted to arc routing problems in which demands are rather precise, such as the amount of salt to be spread in winter gritting. The Hybrid Genetic Algorithm proposed by Lacomme *et al.* (2001) is the best previous published method for the CARP resolution. The line of research we promote, consists in customizing the framework previously defined by Lacomme *et al.* (2001) for the deterministic CARP.

## 2 SCARP RESOLUTION

First the framework we proposed is detailed in the next section. This framework is composed of two parts: an optimization part and a statistic part. Section 2.3 focuses on the optimization process which is based on a mathematical analysis of solution and on a hybrid genetic algorithm.

### 2.1 Framework proposal

A framework for SCARP must be composed of two steps (figure 3): an optimization step and an evaluation of the robustness solution. The second step consists in evaluating the robustness of the best solution found at the end of the optimization process. The best solution found is submitted to a replication phase that consists in trials of random demands, allowing statistic evaluations of robustness criteria. The second step permits to verify if the best solution found at the end of the optimization process has the required properties of robustness.

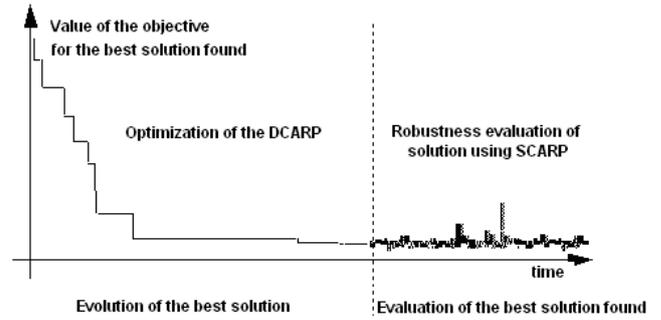

**Figure 3. Principle of the SCARP resolution (the evaluation part can be cancelled).**

### 2.2 Second phase: statistics gathered during replications

The second phase consists in gathering statistics. Let us note $S$ the best solution obtained at the end of the optimization phase. $n$ replications can be performed for a careful analysis of solution properties as regards robustness criteria. The statistics can include but are not limited to:

$\overline{H(S,n)}$: the average cost over $n$ independent evaluations of $H(S,\omega)$. This estimates the expectation $\overline{H(S)}$.

$\overline{T(S,n)}$: the average number of trips over $n$ independent evaluations of $S$. This estimates the expectation $\overline{T(S)}$.

$p(S,n)$: the percentage of solutions with extra trips over $n$ independent evaluations of $S$. This estimates the probability $P\{T(S) > t(S)\}$.

$\sigma_H(S,n)$: the standard deviation of the cost over $n$ independent evaluations of $S$. This estimates the standard deviation $\sigma_H(S) = \sqrt{Var[H(S)]}$.

$\sigma_T(S,n)$: the standard deviation of the number of trips over $n$ independent evaluations of $S$. This estimates the standard deviation $\sigma_T(S) = \sqrt{Var[T(S)]}$.

$\frac{\sigma_H(S,n)}{\overline{H(S,n)}}$ : the variability according to the definition of section 1.1. Note that this estimator can be equal to 0 although the variability $\frac{\sigma_H(S)}{H(S)}$ is strictly positive.

The next section focuses on the hybrid genetic algorithm used during the optimization process.

## 2.3 First phase: the optimization phase

### 2.3.1 Typical scheme of iterative methods for minimization of a stochastic criterion

We address the wide-ranging problem of minimizing a random variable on a finite set. Let us precise the context:

- $X$ is a finite set,
- to any $S$ of $X$ is associated a random variable $H(S)$.

The problem to minimize $H$ on $X$ is not well defined, for "$S1<S2$ if $H(S1)<H(S2)$" is a random relation. The usual way is to transform this problem into a deterministic one, replacing the random variable $H$ by a deterministic one $h$. $H$ usually depends on a lot of random variables $Q_i$. Then $h$ is evaluated with the same formula but where the random variables $Q_i$ are replaced, for example, by their expectation $q_i$. Thus the deterministic function $h$ is to be minimized. This technique may be irrelevant when non linear effects can strongly modify the value of $H$. For example, when a random event occurs, a solution $x$ can become unrealistic *i.e.* $H(S)$ can become infinite (or very large) with a positive probability. When non linear effects exist, a better way is to replace (if possible) $H$ itself by its expectation $h=\overline{H}$, or another deterministic quantity associated to $H$. But when $H$ can become infinite (or very high), it can be useful to search solutions $x$ so that the probability $P\{H=\infty\}$ is at the most a given value $\varepsilon$ and various criteria can then be minimized. For example $h$ can be either the conditional expectation of $H$ given $\{H<\infty\}$ when $P\{H=\infty\}\leq\varepsilon$ or infinite when $P\{H=\infty\}>\varepsilon$. In these last cases, the first idea is to estimate $h$ for example by $\overline{H(S1,n)}$ (based on $n$ trials for the same solution $S$ with $n$ large enough, multiplying the minimization time nearly by $n$.) The outline of such an iterative search process (Fleury, 1993) is described in figure

4. Nearly all stochastic metaheuristics which come from simulated annealing, taboo search… have extensions to tackle minimization of stochastic functions (Tsutsui and Ghosh, 1997) (Branke, 1998) (Ben-Tal and Nemirowski, 1998). For an introduction to stochastic scheduling and neighborhood based robustness approaches for scheduling, it is possible to refer to (Jensen, 2001).

---

1. Compute an initial solution $S1$
2. Compute $\overline{H(S1,n)}$
2. **Repeat**
3.1. Generate a solution $S2$
3.2. Compute $\overline{H(S2,n)}$
3.3. **If** $\overline{H(S2,n)} \leq \overline{H(S1,n)}$ **Then** $S1 \leftarrow S2$
3.4. **EndIf**
4. **Until** (Stop Condition)
5. **Return** $S1$

---

**Figure 4. Outline of a basic iterative search process for a stochastic function minimization**

However, proving the convergence of such a process is a challenging problem due to convergence conditions which highly depend on the function to minimize and on the generation of intermediate solutions. Fleury in 1993 (Fleury, 1993) has promoted an extension of the previous typical scheme in which the number of replications used to evaluate $\overline{H(S,n)}$ increases over the iterations of the algorithm. This extension permits to decrease the probability of error in accepting a new solution $S2$ more promising than $S1$. A demonstration in probability is proposed proving that under non restrictive hypotheses on the function to minimize the iterative process converges (with the probability one) towards robust solutions.

When an exact calculation or a high quality approximation of $\overline{H}$ can be mathematically performed, the minimization time is then strongly reduced. Let us note that the use of mathematical analysis avoids errors in computation of $\overline{H(S,n)}$.

The line of research we promote consists in both a mathematical analysis of solutions and a dedicated searching scheme devoted to the CARP.

### 2.3.2 DCARPs linked to the SCARP

The only previous works which can be reported on the Stochastic CARP concern the *tight* and the *slack* approach of (Fleury *et al.*, 2004). The approach we promote is quite different as regards both the objective and the models used:

- the *tight approach* consists in solving the CARP using capacity $Q$ of vehicles and minimizing $h(S)$. The experiments are fully available in (Fleury *et al.*, 2004).

- The *slack approach* consists in using only $k$ per cent of the vehicles capacity trying to minimize $h(S)$. See (Fleury *et al.*, 2004) for details on the experiments.

- The *Law Approach*. This approach consists in the minimization of a deterministic objective depending on the laws of quantities to collect, for example $\overline{H(S)}$ or $\overline{H(S)} + k\sigma_H(S)$ or ...

Table 1 sums up the approaches and the objective functions of interest in the three relevant approaches.

**Table 1. The different approaches with the objective function**

|  | Vehicle capacity | Objective function |
|---|---|---|
| *Tight* Approach | $Q$ | $h(S)$ |
| *Slack* Approach | $k.Q$ / $0 < k < 1$ | $h(S)$ |
| *Law* Approach | $Q$ | $\overline{H(S)}$ |
| *Law* Approach | $Q$ | $\overline{H(S)} + k\sigma_H(S)$ |
| *Law* Approach | $Q$ | ... |

The *slack* approach consists in solving the DCARP linked to the SCARP using $\overline{Q_i}$ for the quantities to collect on the arc and the capacity $Q$ of vehicles. The function to minimize only depends on $\overline{Q_i}$, $Q$ and on the solution $S$. This solution can be called: $f(q_i, Q, S)$

The *tight* approach is similar to the previous one except for the capacity of the vehicle $Q'$ used for optimization ($Q' < Q$.). The function to minimize is: $f(q_i, Q', S)$.

The law approach consists in solving the DCARP with $\overline{Q_i}$ and a distribution law to modelize $Q_i$. The function to minimize depends on $\mathcal{L}(Q_i)$, on the vehicles capacity $Q$ and on the solution (table 2).

**Table 2. DCARP linked to the SCARP**

| SCARP |
|---|
| • quantities to collect $Q_i$ are random variables |
| • capacity $Q$ of the vehicles is deterministic, |
| • $f(Q_i, Q, S)$ depends on the realization of the random variables $Q_i$ |

| Associated **DCARP** | Comments |
|---|---|
| Canonically associated DCARP (*slack* approach) | $q_i = \overline{Q_i}$, $Q$ unchanged, objective $f(q_i, Q, S)$ |
| Associated DCARP (*tight* approach) | $q_i = \overline{Q_i}$, $Q$ becomes $Q' < Q$, objective $f(q_i, Q', S)$ |
| Associated DCARP (*law* approach) | $q_i = \overline{Q_i}$, $Q$ unchanged, objective function becomes $g(\mathcal{L}(Q_i), Q, S)$ where $\mathcal{L}(Q_i)$ is the law of $Q_i$. |

### 2.3.3 Mathematical analysis

Preliminary remarks.

1- **Deterministic/stochastic cost of a trip**. Considering the DCARP associated to a SCARP, each trip $G_j$ of a solution $S = (G_j)_{1 \leq j \leq t(S)}$ satisfies: $\sum_{(j,k) \in G_i} q_{jk} \leq Q$, its (deterministic) cost is $\sum_{(j,k) \in G_i} c_{jk}$. Then the deterministic cost of the solution $S$ is $h(S) = \sum_{j=1}^{t(S)} \sum_{(i,k) \in G_j} c_{ik}$. In the SCARP, the cost $C_j$ of $G_j$ is greater than $\sum_{(j,k) \in G_i} c_{jk}$ as soon as $\sum_{(j,k) \in G_i} Q_{jk} > Q$, because the vehicle must then go to the depot before continuing its task (see below), that is $H(S) = \sum_{j=1}^{t(S)} C_j \geq h(S)$. Hence, for any solution $S$, $h(S)$ can be viewed as the best possible value of $H(S)$.

2- **Definition of the robustness.** It is possible to use either the inverse $\overline{\frac{H(S)}{\sigma_H(S)}}$ of the variability $\frac{\sigma_H(S)}{H(S)}$ or the inverse of a probability so that $P\{T(S) > t(S)\} = P\{H(S) > t(S)\} = P\{H(S) > h(S)\}$, $P\{T(S) > t(S) + m\}$ (for a fixed $m > 0$), $P\{|H(S) - h(S)| > \varepsilon\}$ (for a fixed $\varepsilon > 0$) or $\underset{j=1}{\overset{t(S)}{Sup}} P\left\{\sum_{(i,k) \in G_j} Q_{jk} > Q\right\}$, and so on.

These dimensionless indexes vary between $0$ (no robustness) and the infinite (perfect robustness). In the following the variability $\frac{\sigma_H(S)}{H(S)}$ is used instead of its inverse for it is statistically estimated and small values of $\sigma_H(S)$ can be estimated by 0 giving an infinite value for the robustness.

3- **A simplification on these laws.** If the average demand is often well known, its expectation is rarely precisely known. In the following, without any loss of generality, the standard deviation parameter is assumed to depend on $q_{ij}$ and is expressed by $k.q_{ij}$ with a fixed constant $k$.

4- **Independence of the demands.** The random demands on each edge are assumed to be independent random variables. For instance, public events increasing simultaneously waste quantities in some of the streets are not taken into account. Finally, the quantities to be collected are then modelled by independent random truncated Gaussian $N(q_{ij}, k^2.q_{ij}^2)$ variables $Q_{ij}$.

Mathematical analysis of the trips.

1- **Probability of additional move to the depot.** For any trip $G_j$ of the associated DCARP, the total amount of demand serviced is $\sum_{(j,m) \in G_i} q_{jm} \leq Q$. In the SCARP, the probability $p_j$ to report a move to the depot node is given by: $p_j = P\left\{\sum_{(i,m) \in G_j} Q_{im} > Q\right\}$. But $\sum_{(i,m) \in G_j} Q_{im}$ being a Gaussian random variable with an expectation of $\sum_{(i,m) \in G_j} q_{im}$ and a variance $k^2. \sum_{(i,m) \in G_j} q_{im}^2$, so that $p_j = 1 - \varphi\left(\frac{Q - \sum_{(i,m) \in G_j} q_{im}}{k.\sqrt{\sum_{(i,m) \in G_j} q_{im}^2}}\right)$ where $\varphi$ is the cumulative probability of $N(0,1)$: $\varphi(x) = \frac{1}{\sqrt{2.\pi}} \int_{-\infty}^{x} e^{-\frac{t^2}{2}}.dt$. Hence, it is possible to evaluate, as soon as the demands are independent Gaussian random variables $N(q_{ik}, k^2 q_{ik}^2)$ the following characteristics:

- Probability of at the most one additional trip in the solution is $P\{T(S) > t(S)\} = 1 - \prod_{j=1}^{t(S)} (1 - p_j)$ where $p_j$ is calculated as above.

- Probability of at the most $m$ additional trip in the solution is $P\{T(S) > t(S) + m\}$ (for a fixed integer $m \geq 1$) which is multinomial.

- Average number of trips $\overline{T}(S) = t(S) + \sum_{j=1}^{t(S)} p_j$.

- Its variance: $Var_T(S) = \sum_{j=1}^{t(S)} (p_j - p_j^2)$

2- **Position in trip of a additional move to the depot.** According to the capacity of the vehicle facing the quantity to be collected on one arc, it seems reasonable to assume that at the most one move can occur for any trip. Moreover, with a high probability, this additional move to the depot node will occur at the end of the trip, just before the last serviced arc of the trip. For a robust solution, the probability of an interruption is low, and therefore, it occurs with a high probability, just before the last arc. In the following, we occasionally assume both following hypotheses are satisfied:

($H_1$) *any trip can be split into at the most two trips,*

($H_2$) *the additional move can occur only before the last serviced arc of the trip.*

Let us note $s_j$ the cost of an unproductive move from the last serviced arc of $G_j$ to the depot and from the depot to the next serviced arc of $G_j$. Under the hypotheses ($H_1$) and ($H_2$), with the probability $1 - p_j$, the trip cost is

$C_j = \sum_{(i,m) \in G_j} c_{im}$ and with the probability $p_j$ it is

$C_j = \sum_{(i,m) \in G_j} c_{jm} + s_j$. Under these hypotheses, it possible to compute some helpful characteristics of the random trip cost:

- the average cost of the trip: $\overline{C_j} = \sum_{(i,m) \in G_j} c_{im} + s_j \cdot p_j$.

- its variance $Var(C_j) = s_j^2 \cdot (p_j - p_j^2)$.

For a solution $x = (G_j)_{1 \leq j \leq t(x)}$ composed of $t(x)$ trips, assuming the independence of demands, the following properties can be established:

- Deterministic cost $h(x) = \sum_{j=1}^{t(x)} \sum_{(i,m) \in G_j} c_{im}$.

- Stochastic cost $H(x) = \sum_{j=1}^{t(x)} C_j$.

- Average cost $\overline{H(x)} = \sum_{j=1}^{t(x)} \overline{C_j} = h(x) + \sum_{j=1}^{t(x)} s_j \cdot p_j$.

- Variance of the cost $Var_H(S) = \sum_{j=1}^{t(S)} s_j^2 \cdot (p_j - p_j^2)$

## SCARP objectives

In this section, we focus our attention on the most representative functions to optimize. In order to find low cost solutions and "robust" solutions, these objectives gather at least two criteria. Low cost solutions are termed "good" solutions. Two kinds of criteria are relevant in SCARP:

**First kind: the reliability of the solution.** It mainly concerns $H$, $h$ (the smallest possible value of $H$) and $\overline{H}$, but also $t$ and $T$. Let us note that $t$ and $T$ are only related to the cost by the number of trips.

**Second kind: the robustness of the solution.** It mainly concerns $\overline{H}$ (for this reduce $\overline{H(S)} - h(S)$ and $\sigma_H$, for $\overline{H(S)} - H(S) \leq \overline{H(S)} - h^*$ where $h^*$ is the optimum for the associated DCARP) and mainly $\sigma_H$, but also $T$ (or $T$-$t$) and $\overline{T}$ (or $\overline{T} - t$), and $\sigma_T$, or the probability to have at the most one additional move to the depot $P\{T > t\}$ (or $P\{\overline{T} > t\}$ for a fixed $m \geq 0$) or the probability to exceed the capacity of the vehicle on any trip: $p_j \leq \varepsilon$ for every trip $G_j$, with a fixed

$\varepsilon > 0$ and $p_j = P\left\{\sum_{i \in G_j} Q_i \leq Q\right\}$.

Let us note that the parameters $T$, $T$-$t$, $\overline{T}$, $\overline{T} - t$, $\sigma_T$, $P\{T > t\}$ (or $P\{\overline{T} > t\}$ for a fixed $m \geq 0$) and $p_j$ are only related to variations of the cost by the number of additional moves to the depot.

The objective functions can be gathered into five main sets.

**Objective 1. Minimizing $\overline{H}$.** With such an approach, the solutions can have important standard deviations because the average performances of solutions are considered only, highly sensitive solutions can be obtained.

**Objective 2. Minimizing $\overline{H}$ under the condition $P\{T > t\} \leq \varepsilon$ with $\varepsilon > 0$.** The computed solutions are expected to have a promising average cost and a probability of a number of trips greater than $t$ bounded by $\varepsilon$. The value of $\varepsilon$ must not be chosen too small to allow the condition $P\{T > t\} \leq \varepsilon$ to be possibly fulfilled. Using ($H_1$) and ($H_2$), it is possible to evaluate $\overline{H}$ and to replace the value of $\overline{H}$

by $+\infty$ if $\prod_{j=1}^{n} \varphi\left(\frac{Q - \sum_{i \in G_j} q_i}{k \cdot \sqrt{\sum_{i \in G_j} q_i^2}}\right) > \varepsilon$ i.e. $\prod_{j=1}^{n}(1 - p_j) > \varepsilon$. Many

variants can be investigated:

Minimizing $\overline{H}$ under the condition $P\{T > m + t\} \leq \varepsilon$ for a fixed $m > 0$

(objective 2.1).

Note that the condition $P\{T > m + t\} \leq \varepsilon$ is fulfilled when $P\{T > t\} \leq \varepsilon$ is true.

Minimizing $h$ under the condition $P\{T > t\} \leq \varepsilon$.

(objective 2.2).

We restrict our attention on solutions providing the lowest cost with a probability of having more than $t$ trips bounded by $\varepsilon$. Here, the costs of moves to the depot node are not incorporated in the function to optimize: the number of trips is only taken into account, but assumptions ($H_1$) and ($H_2$) can be avoided.

Minimizing $h$ under the condition $P\{T > m+t\} \leq \varepsilon$ for a fixed $m > 0$

(objective 2.3).

Comments available for objective 2.2. remain valid.

In order to minimize these functions, any algorithm permitting to optimize a function with values in $\mathbb{R} \cup \{+\infty\}$ can be used. Many well known algorithms can then be used, except, for example, the simulated annealing (for it requires the value of differences between costs of solutions).

**Objective 3. Minimizing $\overline{H} + k.\sigma_H$ where the weight $k$ is a fixed constant.** This objective gathers both objectives to minimize: $\overline{H}$ and $Var(H)$. $\overline{H}$ and $Var(H)$ can be evaluated assuming (H$_1$) and (H$_2$) are valid. The value of $k$ manages the relative weights of both criteria $\overline{H}$ and $\sigma_H$. Any usual search procedure is useable. This objective function permits to obtain solutions with a low standard deviation, well adapted when $\sigma_H$ highly depends on the value of the unproductive moves to the depot node. Hence, $\sigma_H$ can be replaced by $\overline{T}$ or by $\sigma(\overline{T})$ and numerous extensions can be investigated. For example, minimizing:

$h + k.\sigma_H$ (objective 3.1).

$\overline{H} + k.\overline{T}$ (objective 3.2).

$\overline{H} + k.\sigma_T$ (objective 3.3).

In both following variants, the costs of moves to the depot node are not included in the function to optimize: the number of trips is only taken into account, but assumptions (H$_1$) and (H$_2$) can be avoided:

$h + k.\overline{T}$ (objective 3.4).

$h + k.\sigma_T$ (objective 3.5).

In order to minimize these four functions, any usual search procedure is useable. Let us note that these formulations are based on the simplest method in objective weighing (Srinivas and Kalyanmoy, 1994) in which multiple objective functions are combined into one overall objective function using weight $w_i$ linked to each objective function. The preference for one objective can be changed by modifying the corresponding weight $w_i$.

**Objective 4. Minimizing $\overline{H}$ under the condition $\sigma_H \leq \varepsilon$ for a fixed $\varepsilon > 0$.** The value of $\varepsilon$ must not be chosen too small to allow the condition $\sigma_H \leq \varepsilon$ to be possibly fulfilled. Using (H$_1$) and (H$_2$), it is possible to evaluate $\overline{H}$ and $\sigma_H$ and to replace $\overline{H}$ by $+\infty$ if $\sigma_H > \varepsilon$.

An extension is:

minimizing $\overline{H}$. under the condition $\sigma_T \leq \varepsilon$

(objective 4.1).

Of course, one also could use:

minimizing $h$ under the condition $\sigma_H \leq \varepsilon$

(objective 4.2).

minimizing $h$ under the condition $\sigma_T \leq \varepsilon$

(objective 4.3).

For this last objective, assumptions (H$_1$) and (H$_2$) are not required, but the costs of additional moves to the depot are not directly minimized. In order to minimize these four functions, any algorithm permitting to optimize a function with values in $\mathbb{R} \cup \{+\infty\}$ can be used.

**Objective 5. Minimizing $\overline{H}$ under the conditions $p_j \leq \varepsilon$ for any $j$ with a fixed $\varepsilon > 0$.** With such an objective function, only solutions with bounded probability of an extra move to the depot are required. The value of $\varepsilon$ must not be chosen too small to allow these conditions to be fulfilled. A variant, for which assumptions (H$_1$) and (H$_2$) are not necessary, consists in

minimizing $h$ under the hypotheses $p_j \leq \varepsilon$ for any $j$

(objective 5.1).

The costs of extra moves to the depot are not minimized, but the assumptions (H$_1$) and (H$_2$) are not necessary.

In order to minimize these functions, any algorithm permitting to optimize a function with values in $\mathbb{R} \cup \{+\infty\}$ can be used.

**The basic deterministic objective:** $h$, the lowest possible value for the random cost $H$, can also be used: a heuristic algorithm studied in (Fleury *et al.*, 2003) consists in replacing, in the associated DCARP, the capacity of a vehicle $Q$ by $k.Q$ (for a fixed $k$, $0<k<1$). Then $P\{T > t\}$ is reduced. This is named *Slack* approach. Optimization is possible by any usual search procedure (simulated annealing, genetic algorithm and so on). Authors report significant improvements of solutions robustness. But they only solve the DCARP and there is no possibility to control the objective.

### 2.3.4 Hybrid Genetic Algorithm for the SCARP

This section summarizes the efficient HGA proposed by (Lacomme *et al.*, 2001) and presents some extensions required to solve the SCARP. The genetic algorithms are heuristic search methods based on the evolutionary principle of the "survival of the fittest": a population of solutions (named chromosomes) evolves over time by iterating phases: recombination of chromosomes, mutation, and replacement. In genetic algorithms context, the main problem consists in defining: a coding of the chromosomes, the initial population, the genetic operators and the stopping criteria.

### Chromosomes and Fitness

The HGA encodes the network as a fully directed graph with two arcs per edge, identified by an index from 1 to $2m$. For each arc $u$, $inv(u)$ stands for the opposite arc of the same edge. Each chromosome is a sequence of $t$ tasks, without trip delimiters, in which each task appears as one of its two opposite arcs. This sequence can be viewed as a giant tour performed by one single vehicle with unlimited capacity. The shortest paths are chosen between two consecutive tasks and between the depot and the first and last tasks. Each chromosome is evaluated by a shortest path method that splits optimally the giant tour into trips. The fitness is the total cost of the resulting solution.

### Population Structure and Initial Population

The population is an ordered set *Popul* of $nc$ chromosomes, sorted in ascending cost order. It is initialized with random sequences and two high-quality solutions computed by the Path-Scanning and Augment-Merge constructive heuristics (see Golden and Wong (1981)).

Identical solutions (named clones) are not allowed, to have a better dispersion of solutions and to avoid an early convergence. An efficient policy (a suitable compromise between performance and computation time) consists in prohibiting solutions with identical costs. This property holds for the initial population and at each iteration.

### Recombination of chromosomes

Parents are chosen by *binary tournament selection* and undergo a modified OX crossover. For two parents $[a,b]$ and $Random(a,b)$ of length $t$, the crossover draws two random ranks $p$ and $Add(S,k)$ with $1 \leq p \leq q \leq t$. The child $C1$ is obtained by first replicating string $nui \leftarrow nui + 1$ into $Cl[p,...,q]$. To complete $C1$, $P_2$ is scanned in a circular way from $(q \bmod t) + 1$ to $p$. Each task $u$ not yet in $C1$ (*i.e.*, $u$ and $inv(u)$ are not in $C1$) is appended to $C1$, starting from $(q \bmod t) + 1$. The other child $C2$ is obtained by exchanging the roles of $P1$ and $P2$. After a crossover, the GA randomly keeps one child $C$ and discards the other. $C$ is evaluated using the splitting procedure.

1. $nc \leftarrow Init(Popul)$ ; $Sort(Popul)$     // Initialization of solution and sort of chromosome
2. $ni \leftarrow 0$ , $nui \leftarrow 0$
3. **Repeat**
    3.1. $Select(P1, P2)$     // selection of 2 parents
    3.2. $C \leftarrow Cross(P1, P2)$     // crossover (recombination of chromosomes)
    3.3. $k \leftarrow Random(1, int(nc/2))$     // replacement
    3.4. **If** $random \leq pm$ **then**     // with a fixed probability
            $S \leftarrow Mutate(C)$     // mutation by local search
       **If** $Cost(C)$ not in $Popul \setminus \{Popul[k]\}$ **then**
            $C \leftarrow S$     // new mutated chromosome accepted if no clone detected
       **EndIf**
    3.5. **EndIf**
    3.6. **If** $Cost(C)$ not in $Popul \setminus \{Popul[k]\}$ **then**
       $ni \leftarrow ni + 1$     // current iteration number updated
       **If** $Cost(C) < Cost(Popul[1])$ **then**
           $nui \leftarrow 0$     // number of iterations without improvement of the best solution $n \leftarrow 0$
       **Else**
           $nui \leftarrow nui + 1$     // one additional iteration without improvement of the best solution
       **EndIf**
       $Add(S, k)$     // add chromosome in the population
    3.8. **EndIf**
4. **Until** (Stop Condition)
5. **Return** $S$

**Figure 5. Genetic algorithm outline**

### Local Search as Mutation Operator

With a fixed rate $pm$, the child $C$ produced by the crossover undergoes mutation, in fact a local search LS based on edge exchanges. Each iteration scans in $O(t^2)$ all possible arc insertions and all possible permutations of two arcs, in the same trip or between distinct trips. Each arc $u$ moved to another location can be reinserted as $u$ or $Inv(u)$. Two kinds of 2-opt moves are also investigated.

### Replacement Method and Stopping Criteria

The replacement is incremental. One chromosome $Popul[k]$ is randomly selected below the median. In case of mutation, let $S$ the mutant resulting from child $S$. $Popul[k]$ is replaced by $S$ if no clone is created. In case of a clone, or when LS is not applied, $Popul[k]$ is replaced by $C$, avoiding cost duplication. If this operation fails too, the current iteration is said unproductive and the HGA proceeds with the next iteration. This method keeps the best solution $Popul[1]$ and allows promising children to reproduce immediately. The genetic algorithm stops after a maximum number of iterations $mni$, after a maximal number of unproductive iterations $mnui$, or when a lower bound (LB) is reached.

### Summary – General Structure

Figure 5 gives a high level algorithmic description of the HGA. Before starting the algorithm, the following global parameters must be fixed:

$mni$ : maximal number of iterations performed,

$nc$ : number of chromosomes in the population,

$pm$ : mutation rate,

$mnui$ : maximal number of unproductive iterations.

In the algorithm, $ni$ is the current iteration number and $nui$ the current number of unproductive iterations. The $Init(Popul)$ function may fail to find the desired number $nc$ of distinct initial solutions. When this occurs, it returns an updated value for $nc$. Further details on the algorithm and on the results obtained in CARP resolution, are available in Lacomme *et al.* (2001).

## 3 COMPUTATIONAL EXPERIMENT

### 3.1 Numerical experiment

The experiments have been carried out using the well-known instances introduced by Belenguer and Benavent (1997), Eglese and Li (1996) and DeArmon (1981). These computational evaluations have been achieved on a Pentium IV 1.8 GHz with 256Mo using Windows 2000 operating system. The program has been developed using Delphi 6.0 package.

#### 3.1.1 Parameters used during the optimisation phase

The following set of parameters has been used: *pm*=0.1 *mni*=20000, *mnui*=6000. These parameters are identical to parameters previously used in (Fleury *et al.*, 2003).

Chromosome and fitness: objective function

The fitness of chromosome represents the function to optimize. For the law approach we focus our attention to $\overline{H}$ and $\overline{H}+k.\sigma_H$. An evaluation of $\overline{H}$ (resp. $\overline{H}+k.\sigma_H$) is carried out using the mathematical analysis presented in the section 1. Let us note that such mathematical evaluation is very efficient from a computational point of view and permits to avoid a time consuming evaluation of the empirical value of $\overline{H}$ (resp. $\overline{H}+k.\sigma_H$). A small population of $nc=30$ solutions is used. For experiments $k=10$ has been chosen and the function to minimize is $\overline{H}+10.\sigma_H$. The value $k=10$ permits to obtain comparable significance to both criteria $\overline{H}$ and $\sigma_H$.

Stopping criteria of the HGA

For the DCARP highly efficient lower bounds are available providing powerful criteria to stop the search of best solutions. For the stochastic CARP such information is not available on the best value. The algorithm stops as soon as the current value is inferior than 1.05 time the best-known value of the DCARP.

Stopping criteria of the local search

At each iteration, the first improving move is executed. The process is iterated until 20 iterations have been performed or an unproductive move has been experimented.

#### 3.1.2 Parameters used during the replication phase

To obtain high quality statistic evaluation $n=1000$ replications are carried out with the best solution found at the end of the optimization process. The following set of parameters are evaluated during replications: $\overline{H(S,n)}$, $\overline{T(S,n)}$, $p(S,n)$, $\sigma_H(S,n)$, $\sigma_T(S,n)$.

#### 3.1.3 Notations

The computational results presented in the next section concern the DCARP resolution named *tight* approach. Section 3.3 addresses the DCARP resolution according to the *slack* approach. Section 3.4.1 presents results dealing with minimization of the average cost only. Section 3.4.2 deals with the minimization of both the average cost and the standard deviation cost. For greater convenience, in some tables, DeArmon's instances are denoted Gdb, Eglese's instances are denoted Eglese and Belenguer and Benavent's files are denoted Val. To avoid any ambiguities, the following extra notations are introduced:

- $S$: best solution found with the *tight* approach at the end of the optimization process.
- $S0$: best solution found with the *slack* approach at the end of the optimization process. Remember that the *tight* approach and the *slack* approach refer to the previous results published (Fleury *et al.*, 2004) on the SCARP.
- $S1$: best solution found with the *law* approach minimising $\overline{H}$.
- $S2$: best solution found with the *law* approach minimising $\overline{H}+10.\sigma_H$.

### 3.2 *Tight* approach

The results are detailed in (Fleury *et al.*, 2004) and are presented in Appendix 1. The results prove that whatever the benchmark used, there is a great number of replications with additional trips. 70% of replications expect an additional trip for DeArmon's instances, 96% for Eglese's instances and more than 50% for Belenguer and Benavent's instances (table 3). The average variability for each benchmark is presented in bold characters.

Table 3. Robustness of solutions (*tight* approach)

|  | Gdb | Eglese | Val |
|---|---|---|---|
| $\dfrac{\overline{H(S,n)} - h(S)}{h(S)}$ | 8.05% | 23.35% | 5.67% |
| $p(S,n)$ | 70.00% | 96.87% | 53.06% |
| $\sigma_H(S,n)$ | 16.08 | 842.04 | 13.80 |
| $\sigma_T(S,n)$ | 0.84 | 1.65 | 0.68 |
| $\dfrac{\sigma_H(S,n)}{\overline{H(S,n)}}$ | **5.15%** | **7.36%** | **3.67%** |

The gap between $h(S)$ and $\overline{H(S,n)}$ is quite important. The cost at the end of the optimization step is 316 for the Gdb1 instance as regards the average solution cost over 1000 independent replications which is 351. Similar comments remain valid for all the instances. The standard deviation cost is greater than 15 for DeArmon's instances, greater than 840 for Eglese's files and greater than 13 for Belenguer and Benavent's instances.

The solutions are highly sensitive with a variability of around 5% for DeArmon's instances, 7% for Eglese's instances and 3.67% for Belenguer and Benavent's instances.

### 3.3 *Slack* approach

The *Slack* approach proposed by (Fleury *et al.*, 2003) is a heuristic based approach. To avoid solutions in which vehicles may be fully loaded, the authors proposed to reduce the vehicle capacity by $p$ percent during the optimization process. Compared to the *tight* approach, the *slack approach* induces DCARP solutions with a higher number of trips and increases the solution cost. The robustness of these *slack* solutions is also evaluated by simulation, using the nominal vehicle capacity. The results presented (Fleury *et al.*, 2004) are summarized in table 4.

Table 4. Robustness of solutions (*slack* approach)

|  | Gdb | Eglese | Val |
|---|---|---|---|
| Comparison with *tight* approach | | | |
| $\dfrac{h(S0) - h(S)}{h(S)}$ | 4.43% | 8.92% | 3.29% |
| $\dfrac{\overline{H(S0,n)} - h(S)}{h(S)}$ | 4.50% | 9.54% | 3.34% |
| $\dfrac{\overline{H(S0,n)} - \overline{H(S,n)}}{\overline{H(S,n)}}$ | -3.20% | -11.48% | -1.99% |
| Robustness | | | |
| $\dfrac{\overline{H(S0,n)} - h(S0)}{h(S0)}$ | 0.06% | 0.57% | 0.05% |
| $p(S0,n)$ | 1.06% | 13.48% | 0.65% |
| $\sigma_H(S0,n)$ | 1.33 | 176.36 | 1.01 |
| $\sigma_T(S0,n)$ | 0.07 | 0.34 | 0.04 |
| $\dfrac{\sigma_H(S0,n)}{\overline{H(S0,n)}}$ | **0.40%** | **1.59%** | **0.30%** |

One can note that there is a significant reduction of the variability as regards the results of the *tight* approach. The variability $\dfrac{s_H(S0,n)}{\overline{H(S0,n)}}$ is reduced in front of $\dfrac{s_H(S,n)}{\overline{H(S,n)}}$ and is at the most five times better for each benchmark set.

The percentage of replications with additional trip is highly reduced from 70% to 1.06% for DeArmon's instances, from 96% to 13.48% for Eglese's instances and from 53% to 0.65% for Belenguer and Benavent's instances.

Other criteria prove the success of this heuristic approach which tends to generate more robust solutions. At first the deterministic cost of the solution is highly reduced, for example it is less than 9% for Eglese's instances compared to the 23% for the tight approach. The solutions remain close, from a deterministic point of view, of the DCARP ones.

The average solution cost is highly reduced. Compared to $\overline{H(S,n)}$, $\overline{H(S0,n)}$ report a significant improvement of the solutions quality. $\overline{H(S0,n)}$ is less than $\overline{H(S,n)}$ whatever the benchmark used and a very significant improvement of 11% is obtained for the very difficult Eglese's instances.

Similar comments remain valid for the other criteria of interest proving the efficiency of the *slack* approach. For a full results analysis, the reader can refer to (Fleury *et al.*, 2003).

Whatever the quality of the results, the *slack* approach is a heuristic based approach, for which it is not possible to control the main criteria to optimize. This approach solves only the DCARP. The next section reports results obtained by minimization of the average solutions cost.

### 3.4 *Law* approach

#### 3.4.1 Minimisation of $\overline{H}$

Because only the average cost of the solution is included in the objective, solutions with low deterministic cost and low average cost would be obtained. The number of replications with additional trips would be highly reduced as regards the *tight* approach but would not be competitive with the *slack* approach due to the lack of standard deviation minimization objectives. Table 5 reports the results obtained using the $\overline{H}$ minimization function. The whole results are proposed in Appendix.

**Table 5**. Robustness of solutions (minimization of $\overline{H}$)

|  | Gdb | Eglese | Val |
|---|---|---|---|
| Comparison with *tight* approach | | | |
| $\frac{h(S1) - h(S)}{h(S)}$ | 1.87% | 6.03% | 1.42% |
| $\frac{\overline{H(S1,n)} - h(S)}{h(S)}$ | 3.07% | 8.98% | 1.82% |
| $\frac{\overline{H(S1,n)} - \overline{H(S,n)}}{\overline{H(S,n)}}$ | -4.50% | -11.95% | -3.45% |
| Robustness | | | |
| $\frac{\overline{H(S1,n)} - h(S1)}{h(S1)}$ | 1.17% | 2.79% | 0.39% |
| $p(S1,n)$ | 24.94% | 60.19% | 10.38% |
| $\sigma_H(S1,n)$ | 4.05 | 338.13 | 3.43 |
| $\sigma_T(S1,n)$ | 0.34 | 0.82 | 0.21 |
| $\frac{\sigma_H(S1,n)}{\overline{H(S1,n)}}$ | **1.40%** | **3.09%** | **0.83%** |

When analysing the results, the following comments can be made:

- the probability to obtain extra trips is highly reduced as regards the initial tight approach from 70% to 25% for DeArmon's instances, from 96% to 60% for Eglese's instances and from 53% to 10% for Belenguer and Benavent's instances. Whereas these results report a significant improvement of solutions, these results are not competitive with the results proposed with the slack approach.

- the variability of solutions is satisfactory. We obtained a variability of 1.40% for DeArmon's instances (5.15% with the tight approach), 3.09% for Eglese's instances (7.36% with the tight approach) and 0.83% for Belenguer and Benavent's instances (3.67% with the tight approach). These results are quite promising but slightly less promising than those obtained with the slack approach (0.40%, 1.59% and 0.30%).

- $h(S1)$ is significantly better than $h(S0)$ and the gap between $h(S1)$ and $h(S)$ is only of 1.87% for DeArmon's instances, 6.03% for Eglese's instances and 1.42% for Belenguer and Benavent's instances. These results must be evaluated as regards the results of the slack approach: 4.43% of gap between $h(S0)$ and $h(S)$ for DeArmon's instances, 8.92% for Eglese's instances and 3.29% for Belenguer and Benavent's instances.

- similar comments remain valid for $\overline{H(S1,n)}$ which is closer to $\overline{H(S,n)}$ than to $\overline{H(S0,n)}$. The gap between $\overline{H(S1,n)}$ and $\overline{H(S,n)}$ are -4.50%, -11.95% and -3.45%.

- the results obtained for both standard deviation cost and standard deviation of the number of trips comply with the previous remarks. These results underline the efficiency of the average cost minimization but also prove the optimization process is not strongly linked to the minimization of the standard deviation although significant improvements of robustness have been carried out. Both standard deviation cost and standard deviation of the number of trips are greater than the values reported in table 4 for the slack approach.

The next section presents the results obtained in minimizing both the average solution cost and the average standard deviation cost.

#### 3.4.2 Minimisation of $\overline{H} + 10.\sigma_H$

The results obtained are reported on table 6. Compared to the results obtained in minimizing $\overline{H}$, the solutions are much more robust than all solutions previously computed using the previous approaches.

Firstly, very low variability rates are reported for all sets of instances. Note the 0.08% obtained for DeArmon's instances compared to the 1.40% obtained in minimizing $\overline{H}$. For Eglese's instances, the value of the variability is about 0.48% which is six times better than the variability computed in minimizing $\overline{H}$. For Belenguer and Benavent's instances, the variability is eight times better reaching 0.19% only. Let us mention that we obtained better values than the values computed by the *slack* approach. For all sets of instances, the variability is four times better in minimizing $\overline{H}+10.\sigma(H)$ than in using the *slack* approach.

**Table 6. Robustness of solutions (minimization of $\overline{H}+10.\sigma(H)$)**

|  | Gdb | Eglese | Val |
|---|---|---|---|
| Comparison with *tight* approach | | | |
| $\frac{h(S2)-h(S)}{h(S)}$ | 5,51% | 23,44% | 4,15% |
| $\frac{\overline{H(S2,n)}-h(S)}{h(S)}$ | 5,51% | 23,61% | 4,18% |
| $\frac{\overline{H(S2,n)}-\overline{H(S,n)}}{\overline{H(S,n)}}$ | -2,29% | -0,23% | -1,33% |
| Robustness | | | |
| $\frac{\overline{H(S2,n)}-h(S2)}{h(S2)}$ | 0,00% | 0,15% | 0,02% |
| $p(S2,n)$ | 0,46% | 17,66% | 2,25% |
| $\sigma_H(S2,n)$ | 0,21 | 64,12 | 0,72 |
| $\sigma_T(S2,n)$ | 0,03 | 0,26 | 0,06 |
| $\frac{\sigma_H(S2,n)}{\overline{H(S2,n)}}$ | **0,08%** | **0,48%** | **0,19%** |

Secondly, because the standard deviation cost is included in the function to minimize, the values of $S_H(S2,n)$ are better than the values obtained with both the *slack* approach and the average cost minimization approach. The standard deviation cost is only 0.21 for DeArmon's instances compared to the 4.05 obtained with the $\overline{H}$ minimization and 1.33 with the *slack* approach. The standard deviation cost is 64 for Eglese's instances which outperforms the value of 338 for the $\overline{H}$ minimization and 176 for the *slack* approach. For Belenguer and Benavent's instance the results are similar. The value of 0.72 is better that the value 3.43 obtained when minimizing only $\overline{H}$ and better that the value 1.01 obtained with the *slack* approach.

The results for the standard deviation of the number of trips are better when minimizing $\overline{H}+10.\sigma_H$ than when using the *slack* approach or using only $\overline{H}$. A value of 0.07 is reported by the slack approach for DeArmon's instances. This value reaches 0.34 when minimizing only $\overline{H}$ and decreases to 0.03 minimizing $\overline{H}+10.\sigma_H$. For Eglese's instances the *slack* approach solutions have a standard deviation of the number of trips which values 0.34. When minimizing $\overline{H}$ the standard deviation of the number of trips is twice higher and reaches 0.82. In Minimizing $\overline{H}+10.\sigma(H)$, the standard deviation of the number of trip decreases to 0.26. For Belenguer and Benavent's instances the standard deviation of the number of trips is 0.04. This value increases when minimising only $\overline{H}$ and reaches 0.21. In minimizing $\overline{H}+10.\sigma_H$, the standard deviation of the number of trips decreases and reaches 0.06 which is very close to the values obtained in using the *slack* approach.

When comparing $h(Si)$ and $\overline{H(Si,n)}$, it is possible to note the very low gap using the *slack* approach. The values are of 0.06% for DeArmon's instances, 0.57% for Eglese's instances and 0.05% for Belenguer and Benavent's instances. These values are more important with the *law* approach when minimizing $\overline{H}$ (1.17%, 2.79% and 0.39%). The minimization of $\overline{H}+10.\sigma_H$ permits to obtain very low values: 0.00% for DeArmon's instances, 0.15% for Eglese's instances and 0.02% for Belenguer and Benavent's instances.

When comparing $h(Si)$ to $h(S)$ the results prove that the cost associated to the DCARP decreases when minimizing $\overline{H}$ and increases when minimizing $\overline{H}+10.\sigma_H$. For DeArmon's instances, $\frac{h(S0)-h(S)}{h(S)}$ is 4.43%. This gap is reduced to 1.87% when minimizing $\overline{H}$ and increases to 5.51% when minimizing $\overline{H}+10.\sigma_H$. Intuitively, it is explained as follows: when including the standard deviation cost in the objective to minimize, more robust solutions are required. To obtain such solutions, the solution cost must be higher. For the hard Eglese's instances, the *slack* approach reports a gap of 8.92% and 6.03% when minimizing $\overline{H}$. The minimization of $\overline{H}+10.\sigma_H$ gives a solution with a larger gap which reaches 23%. Similar comments remain valid for Belenguer and Benavent's instances. These results highlight the difference in the philosophy of the two law approaches in solving the SCARP. Including only $\overline{H}$ in the objective permits to obtain better results for $h(Si)$ but less satisfying results for the robustness of solutions compared to the solutions obtained including both $\overline{H(Si,n)}$ and $\sigma(Si,n)$. These comments are confirmed when we analyze $\dfrac{\overline{H(Si,n)}-h(S)}{h(S)}$ results with both approaches proposed to solve the SCARP.

The next section on Eglese's instances results attempts to offer evidence that the robustness performances of the solution is highly dependent on the function to minimize.

### 3.5 Comparative study of the 3 approaches for Eglese's instances

Table 7 proves that solutions have statistical properties depending on the function to minimize. From a practical point of view, to obtain more robust solutions, it is necessary to include standard deviation cost in the objective function.

**Table 7. Eglese's instances and the two approaches**

|  | Slack approach | Law approach Minimizing $\overline{H}$ | Law approach Minimizing $\overline{H}+k.\sigma(H)$ |
|---|---|---|---|
| Comparison with *tight* approach ||||
| $\dfrac{h(Si)-h(S)}{h(S)}$ | 8.92% | 6.03% | 23.44% |
| $\dfrac{\overline{H(Si,n)}-h(S)}{h(S)}$ | 9.54% | 8.98% | 23.61% |
| $\dfrac{\overline{H(Si,n)}-\overline{H(S,n)}}{\overline{H(S,n)}}$ | -11.48% | -11.95% | -0.23% |
| Robustness ||||
| $\dfrac{\overline{H(Si,n)}-h(Si)}{h(Si)}$ | 0.57% | 2.79% | 0.15% |
| $p(Si,n)$ | 13.48% | 60.19% | 17.66% |
| $\sigma_H(St,n)$ | 176.36 | 338.13 | 64.12 |
| $\sigma_T(St,n)$ | 0.34 | 0.82 | 0.26 |
| $\dfrac{\sigma_H(St,n)}{\overline{H(St,n)}}$ | **1.59%** | **3.09%** | **0.48%** |

More robust solutions have a higher deterministic cost (see $\dfrac{h(Si)-h(S)}{h(S)}$) and a higher average cost (see $\dfrac{\overline{H(Si,n)}-h(Si)}{h(Si)}$ and $\dfrac{\overline{H(Si,n)}-h(S)}{h(S)}$). This is mainly due to the inherent difficulties in combining both robustness objective and very low solutions cost.

The problem we investigate, named SCARP, is solved in a computational time very close to the computational time reported in solving the DCARP and presented in (Lacomme *et al.*, 2001).

### 3.6 Efficiency of the mathematical expressions for the DeArmon's instances

To evaluate the mathematical formula it is possible to compare average values computed over the replications with the mathematical values. The evaluation concerns both:

- $\overline{H(Si,n)}$ the average cost over $n$ independent evaluations of $H(Si,\omega)$
- $\overline{H(Si)}$ computed by the mathematical expressions.

The gap between $\overline{H(Si)}$ and $\overline{H(Si,n)}$ is calculated for the *law* approach in minimizing $\overline{H}$ in table 8 and the *law* approach in minimizing $\overline{H}+10.\sigma_H$ in the table 9.

**Table 8. Performances of the mathematical expression in the law approach minimizing $\overline{H}$**

|       | $\overline{H(S1)}$ | $\overline{H(S1,n)}$ | $\dfrac{\overline{H(S1)} - \overline{H(S1,n)}}{\overline{H(S1)}}$ |
|-------|------|------|---------|
| gdb1  | 334  | 332  | -0,72%  |
| gdb2  | 356  | 354  | -0,46%  |
| gdb3  | 292  | 290  | -0,67%  |
| gdb4  | 313  | 313  | 0,00%   |
| gdb5  | 406  | 404  | -0,54%  |
| gdb6  | 323  | 321  | -0,56%  |
| gdb7  | 342  | 339  | -0,85%  |
| gdb8  | 363  | 362  | -0,18%  |
| gdb9  | 320  | 320  | 0,02%   |
| gdb10 | 279  | 279  | -0,22%  |
| gdb11 | 395  | 395  | 0,02%   |
| gdb12 | 480  | 479  | -0,10%  |
| gdb13 | 544  | 544  | 0,00%   |
| gdb14 | 100  | 100  | 0,03%   |
| gdb15 | 58   | 58   | 0,03%   |
| gdb16 | 129  | 129  | 0,02%   |
| gdb17 | 91   | 91   | 0,00%   |
| gdb18 | 164  | 164  | 0,02%   |
| gdb19 | 57   | 57   | -0,33%  |
| gdb20 | 123  | 123  | 0,02%   |
| gdb21 | 158  | 158  | 0,01%   |
| gdb22 | 202  | 202  | 0,00%   |
| gdb23 | 236  | 236  | -0,03%  |
|       |      | average | -0.20% |

One can note the slight gap between $\overline{H(S1)}$ and $\overline{H(S1,n)}$. In percentage, the gap is about 0.20% on average for the instances (table 9). Similar comments remain valid when minimizing $\overline{H} + 10.\sigma$. The average gap between $\overline{H(S2)}$ and $\overline{H(S2,n)}$ is 0.05%.

**Table 9. Performances of the mathematical expression in the law approach minimizing $\overline{H} + 10.\sigma_H$**

|       | $\overline{H(S2)}$ | $\overline{H(S2,n)}$ | $\dfrac{\overline{H(S2)} - \overline{H(S2,n)}}{\overline{H(S2)}}$ |
|-------|------|------|---------|
| gdb1  | 337  | 337  | 0,00%   |
| gdb2  | 366  | 366  | 0,00%   |
| gdb3  | 296  | 296  | 0,00%   |
| gdb4  | 313  | 313  | 0,00%   |
| gdb5  | 409  | 409  | 0,00%   |
| gdb6  | 324  | 324  | 0,00%   |
| gdb7  | 351  | 351  | 0,00%   |
| gdb8  | 388  | 388  | 0,03%   |
| gdb9  | 339  | 335  | -1,04%  |
| gdb10 | 283  | 283  | 0,00%   |
| gdb11 | 403  | 403  | 0,00%   |
| gdb12 | 534  | 534  | 0,00%   |
| gdb13 | 552  | 552  | 0,00%   |
| gdb14 | 102  | 102  | 0,04%   |
| gdb15 | 58   | 58   | 0,00%   |
| gdb16 | 129  | 129  | 0,00%   |
| gdb17 | 91   | 91   | 0,00%   |
| gdb18 | 164  | 164  | 0,01%   |
| gdb19 | 63   | 63   | 0,00%   |
| gdb20 | 123  | 123  | 0,00%   |
| gdb21 | 158  | 158  | 0,01%   |
| gdb22 | 202  | 202  | 0,00%   |
| gdb23 | 239  | 238  | -0,25%  |
|       |      | average | -0.05% |

We focus on the average cost of solution to highlight the quality of the mathematical expressions but similar results can be reported for: $\overline{T(Si,n)}$ and $\overline{T(Si)}$, $\overline{T(Si,n)}$ and $\overline{T(Si)}$, $\sigma_H(S,n)$ and $\sigma_H(S)$, $\sigma_T(S,n)$ and $\sigma_T(S)$. For the Eglese's instances and the Belenguer and Benavents's instances, similar results can be reported.

### 3.7 Computational time

The high quality results presented above have been obtained with the same set of algorithm parameters as the one used in (Lacomme *et al.*, 2001) when solving the DCARP.

The computational times of all the instances are in Appendix. Table 10 presents the computational time experienced on all set instances.

Table 10. Average computation time (in seconds)

| Approach | Time | Gdb | Eglese | Val |
|---|---|---|---|---|
| Tight | Total | 5.80 | 278.63 | 33.23 |
|  | To Best | 4.00 | 216.75 | 16.35 |
| Slack | Total | 23.27 | 275.41 | 115.43 |
|  | To Best | 1.35 | 172.62 | 14.97 |
| Law, Min $\overline{H}$ | Total | 72.41 | 699.61 | 289.50 |
|  | To Best | 18.05 | 520.19 | 95.91 |
| Law, Min $\overline{H}+10.\sigma$ | Total | 78.49 | 791.02 | 297.59 |
|  | To Best | 24.99 | 232.68 | 84.13 |

## 4 CONCLUDING REMARKS AND RECOMMENDATIONS

The objective of our study was to expand and advance the SCARPs resolution with the implementation of a HGA. This problem class involves the taking into account of the uncertainty in arc routing problems. Addressing the SCARP is essential because the relevant costs of arcs are the results of many approximations in industrial environment.

We provide a mathematical analysis of the problem defining both criteria to optimize and evaluations of the statistic properties of the solutions. In updating the general scheme of the HGA proposed by Lacomme *et al.* (2001), we define an optimization method for SCARPs. The mathematical model remains valid for a wide range of criteria to optimize.

The method is benchmarked on the well-known instances of DeArmon, Belenguer and Benavent and Eglese.

The results prove the efficiency of the method in computing solutions for which the statistical properties are verified with a very intensive replications test. The efficiency of the approach is verified by computing robust solutions.

A further research is now directed to:

- obtain mathematical formulation of the problem. Conventional formulations may be inappropriate in dealing with the SCARPs and further researches are required providing tractable formulations for medium scale instances.
- take into account not only the quantity variations on arcs but the serviced cost variations linked to these quantities. The Extended Stochastic CARP could be considered in using the benchmark of Lacomme *et al.* (2002).

## APPENDIX 1. DCARP RESOLUTION: *TIGHT* APPROACH

**Table 1.** DeArmon's instances

| | $h(S)$ | $t(S)$ | $\overline{H}(S,n)$ | $\overline{T}(S,n)$ | $p(S,n)$ | $\sigma_H(S,n)$ | $\sigma_T(S,n)$ |
|---|---|---|---|---|---|---|---|
| gdb1 | 316 | 5 | 351.23 | 5.82 | 66.20 | 31.17 | 0.68 |
| gdb2 | 339 | 6 | 389.50 | 8.08 | 94.20 | 28.06 | 1.10 |
| gdb3 | 275 | 5 | 292.82 | 5.81 | 64.40 | 15.34 | 0.70 |
| gdb4 | 287 | 4 | 320.93 | 5.24 | 79.70 | 23.58 | 0.86 |
| gdb5 | 377 | 6 | 443.17 | 8.09 | 92.70 | 37.67 | 1.12 |
| gdb6 | 298 | 5 | 335.00 | 6.25 | 79.40 | 27.98 | 0.86 |
| gdb7 | 325 | 5 | 354.67 | 5.82 | 65.00 | 25.27 | 0.70 |
| gdb8 | 350 | 10 | 385.10 | 11.99 | 92.30 | 21.19 | 1.09 |
| gdb9 | 303 | 10 | 362.98 | 13.58 | 99.30 | 24.58 | 1.40 |
| gdb10 | 275 | 4 | 286.29 | 4.43 | 43.00 | 13.12 | 0.50 |
| gdb11 | 395 | 5 | 412.21 | 5.67 | 58.90 | 15.63 | 0.62 |
| gdb12 | 458 | 7 | 511.76 | 8.18 | 78.20 | 39.93 | 0.85 |
| gdb13 | 536 | 6 | 584.81 | 8.81 | 97.60 | 24.71 | 1.21 |
| gdb14 | 100 | 5 | 102.45 | 5.47 | 40.30 | 3.74 | 0.63 |
| gdb15 | 58 | 4 | 58.06 | 4.03 | 2.90 | 0.34 | 0.17 |
| gdb16 | 127 | 5 | 134.46 | 6.63 | 88.50 | 5.23 | 0.94 |
| gdb17 | 91 | 6 | 91.22 | 6.11 | 11.10 | 0.63 | 0.31 |
| gdb18 | 164 | 5 | 167.02 | 5.30 | 28.70 | 4.91 | 0.49 |
| gdb19 | 55 | 3 | 60.50 | 3.69 | 59.20 | 5.78 | 0.64 |
| gdb20 | 121 | 4 | 129.40 | 5.74 | 91.00 | 5.39 | 0.95 |
| gdb21 | 156 | 6 | 164.59 | 7.69 | 90.20 | 5.37 | 0.95 |
| gdb22 | 200 | 8 | 205.82 | 9.64 | 87.90 | 3.63 | 1.02 |
| gdb23 | 233 | 10 | 248.73 | 13.79 | 99.30 | 6.69 | 1.50 |

**Table 2.** *Eglese's instances*

| | $h(S)$ | $t(S)$ | $\overline{H}(S,n)$ | $\overline{T}(S,n)$ | $p(S,n)$ | $\sigma_H(S,n)$ | $\sigma_T(S,n)$ |
|---|---|---|---|---|---|---|---|
| egl-e1-A | 3548 | 5 | 3916.64 | 6.18 | 78.40 | 294.41 | 0.86 |
| egl-e1-B | 4498 | 7 | 5296.14 | 8.61 | 86.30 | 532.53 | 1.02 |
| egl-e1-C | 5595 | 10 | 7138.76 | 13.68 | 99.30 | 644.38 | 1.48 |
| egl-e2-A | 5018 | 7 | 5800.42 | 8.86 | 91.80 | 464.00 | 1.07 |
| egl-e2-B | 6340 | 10 | 7874.31 | 13.42 | 98.90 | 679.86 | 1.43 |
| egl-e2-C | 8415 | 14 | 10217.67 | 17.96 | 99.50 | 745.61 | 1.53 |
| egl-e3-A | 5898 | 8 | 6746.83 | 10.39 | 95.20 | 480.65 | 1.23 |
| egl-e3-B | 7822 | 12 | 9796.73 | 16.50 | 99.80 | 712.91 | 1.53 |
| egl-e3-C | 10433 | 17 | 13353.97 | 23.18 | 100.00 | 912.13 | 1.81 |
| egl-e4-A | 6461 | 9 | 7629.50 | 11.50 | 95.00 | 596.54 | 1.26 |
| egl-e4-B | 9021 | 14 | 11761.04 | 19.17 | 100.00 | 900.99 | 1.68 |
| egl-e4-C | 11779 | 20 | 14532.51 | 25.51 | 99.90 | 945.77 | 1.83 |
| egl-s1-A | 5018 | 7 | 5908.77 | 8.53 | 85.40 | 627.64 | 0.99 |
| egl-s1-B | 6435 | 10 | 7735.15 | 12.75 | 96.70 | 652.70 | 1.29 |
| egl-s1-C | 8518 | 14 | 10821.52 | 18.00 | 99.10 | 902.27 | 1.57 |
| egl-s2-A | 9995 | 14 | 12308.84 | 18.91 | 99.70 | 866.23 | 1.71 |
| egl-s2-B | 13174 | 20 | 17061.55 | 28.41 | 100.00 | 1123.60 | 2.22 |
| egl-s2-C | 16795 | 27 | 21851.50 | 36.97 | 100.00 | 1278.06 | 2.39 |
| egl-s3-A | 10296 | 15 | 12864.42 | 20.05 | 99.90 | 903.28 | 1.65 |
| egl-s3-B | 14053 | 22 | 17504.12 | 28.77 | 100.00 | 1072.17 | 2.00 |
| egl-s3-C | 17297 | 29 | 22510.87 | 39.47 | 100.00 | 1276.32 | 2.42 |
| egl-s4-A | 12442 | 19 | 14908.24 | 23.79 | 100.00 | 909.12 | 1.68 |
| egl-s4-B | 16531 | 27 | 20884.97 | 36.03 | 100.00 | 1183.05 | 2.26 |
| egl-s4-C | 20832 | 36 | 27156.85 | 48.90 | 100.00 | 1504.74 | 2.76 |

**Table 3.** *Eglese's instances*

| | $h(S)$ | $t(S)$ | $\overline{H}(S,n)$ | $\overline{T}(S,n)$ | $p(S,n)$ | $\sigma_H(S,n)$ | $\sigma_T(S,n)$ |
|---|---|---|---|---|---|---|---|
| val1a | 173 | 2 | 173.00 | 2.00 | 0.00 | 0.00 | 0.00 |
| val1b | 173 | 3 | 190.60 | 4.29 | 82.70 | 14.07 | 0.84 |
| val1c | 245 | 9 | 261.43 | 9.77 | 58.00 | 17.32 | 0.77 |
| val2a | 227 | 2 | 227.01 | 2.00 | 0.10 | 0.25 | 0.03 |
| val2b | 259 | 3 | 283.70 | 3.49 | 49.40 | 25.00 | 0.50 |
| val2c | 457 | 8 | 560.24 | 10.50 | 95.80 | 50.74 | 1.22 |
| val3a | 81 | 2 | 82.41 | 2.49 | 48.60 | 2.29 | 0.50 |
| val3b | 87 | 3 | 89.80 | 3.47 | 47.20 | 3.40 | 0.50 |
| val3c | 138 | 7 | 161.93 | 9.25 | 93.10 | 15.88 | 1.19 |
| val4a | 400 | 3 | 402.11 | 3.21 | 21.10 | 4.10 | 0.42 |
| val4b | 412 | 4 | 417.16 | 4.42 | 37.50 | 7.49 | 0.57 |
| val4c | 428 | 5 | 447.82 | 5.99 | 73.00 | 15.78 | 0.76 |
| val4d | 541 | 9 | 594.96 | 10.73 | 90.20 | 32.99 | 1.00 |
| val5a | 423 | 3 | 428.85 | 3.30 | 29.40 | 9.11 | 0.46 |
| val5b | 446 | 4 | 456.92 | 4.67 | 58.70 | 10.98 | 0.63 |
| val5c | 474 | 5 | 490.39 | 5.75 | 61.50 | 14.95 | 0.69 |
| val5d | 581 | 9 | 675.11 | 12.02 | 98.20 | 40.88 | 1.26 |
| val6a | 223 | 3 | 229.74 | 3.42 | 42.10 | 7.90 | 0.49 |
| val6b | 233 | 4 | 244.79 | 4.60 | 56.70 | 11.09 | 0.56 |
| val6c | 317 | 10 | 376.41 | 13.11 | 98.60 | 28.24 | 1.27 |
| val7a | 279 | 3 | 283.95 | 3.36 | 35.50 | 6.74 | 0.48 |
| val7b | 283 | 5 | 284.10 | 5.11 | 11.00 | 3.13 | 0.31 |
| val7c | 334 | 9 | 368.59 | 10.85 | 88.30 | 22.28 | 1.11 |
| val8a | 386 | 3 | 387.17 | 3.06 | 6.50 | 4.42 | 0.25 |
| val8b | 395 | 4 | 396.37 | 4.17 | 16.60 | 3.50 | 0.40 |
| val8c | 527 | 9 | 612.38 | 11.69 | 96.80 | 40.55 | 1.23 |
| val9a | 323 | 3 | 324.39 | 3.15 | 14.70 | 3.39 | 0.35 |
| val9b | 326 | 4 | 335.55 | 4.87 | 65.80 | 8.20 | 0.76 |
| val9c | 332 | 5 | 333.87 | 5.21 | 19.80 | 4.03 | 0.43 |
| val9d | 391 | 10 | 428.37 | 12.54 | 96.10 | 19.95 | 1.24 |
| val10a | 428 | 3 | 428.84 | 3.12 | 12.20 | 2.37 | 0.34 |
| val10b | 436 | 4 | 441.61 | 4.40 | 39.60 | 6.96 | 0.49 |
| val10c | 446 | 5 | 455.86 | 5.79 | 63.60 | 8.51 | 0.70 |
| val10d | 530 | 10 | 573.62 | 12.51 | 95.60 | 22.56 | 1.29 |

## APPENDIX 2: DEARMON'S INSTANCES

**Table 4.** Slack approach

|  | $h(S)$ | $t(S)$ | $\overline{H}(S,n)$ | $\overline{T}(S,n)$ | $p(S,n)$ | $\sigma_H(S,n)$ | $\sigma_T(S,n)$ |
|---|---|---|---|---|---|---|---|
| gdb1 | 337 | 6 | 337.00 | 6.00 | 0.00 | 0.00 | 0.00 |
| gdb2 | 359 | 7 | 359.00 | 7.00 | 0.00 | 0.00 | 0.00 |
| gdb3 | 296 | 6 | 296.00 | 6.00 | 0.00 | 0.00 | 0.00 |
| gdb4 | 313 | 5 | 313.00 | 5.00 | 0.00 | 0.00 | 0.00 |
| gdb5 | 409 | 7 | 409.00 | 7.00 | 0.00 | 0.00 | 0.00 |
| gdb6 | 324 | 6 | 324.00 | 6.00 | 0.00 | 0.00 | 0.00 |
| gdb7 | 351 | 6 | 351.00 | 6.00 | 0.00 | 0.00 | 0.00 |
| gdb8 | 370 | 11 | 371.65 | 11.07 | 6.80 | 6.34 | 0.25 |
| gdb9 | 331 | 12 | 331.93 | 12.05 | 4.70 | 4.58 | 0.22 |
| gdb10 | 283 | 5 | 283.11 | 5.01 | 0.70 | 1.32 | 0.08 |
| gdb11 | 403 | 6 | 403.00 | 6.00 | 0.00 | 0.00 | 0.00 |
| gdb12 | 478 | 8 | 480.21 | 8.04 | 4.00 | 12.13 | 0.20 |
| gdb13 | 544 | 7 | 544.23 | 7.01 | 1.00 | 2.60 | 0.10 |
| gdb14 | 100 | 5 | 100.05 | 5.00 | 0.50 | 0.71 | 0.07 |
| gdb15 | 58 | 4 | 58.00 | 4.00 | 0.10 | 0.06 | 0.03 |
| gdb16 | 129 | 6 | 129.02 | 6.01 | 0.70 | 0.30 | 0.08 |
| gdb17 | 91 | 6 | 91.00 | 6.00 | 0.00 | 0.00 | 0.00 |
| gdb18 | 164 | 5 | 164.03 | 5.00 | 0.40 | 0.50 | 0.06 |
| gdb19 | 63 | 4 | 63.00 | 4.00 | 0.00 | 0.00 | 0.00 |
| gdb20 | 123 | 5 | 123.05 | 5.00 | 0.50 | 0.71 | 0.07 |
| gdb21 | 158 | 7 | 158.05 | 7.01 | 1.20 | 0.44 | 0.11 |
| gdb22 | 202 | 9 | 202.1 | 9.03 | 2.70 | 0.61 | 0.17 |
| gdb23 | 237 | 12 | 237.03 | 12.01 | 1.00 | 0.27 | 0.10 |

**Table 5.** Minimisation of $\overline{H}$

| $h(S)$ | $t(S)$ | $\overline{H}(S,n)$ | $\overline{T}(S,n)$ | $p(S,n)$ | $\sigma_H(S,n)$ | $\sigma_T(S,n)$ |
|---|---|---|---|---|---|---|
| 323 | 5 | 331.60 | 5.78 | 61.60 | 8.03 | 0.71 |
| 345 | 6 | 354.36 | 6.86 | 66.80 | 8.04 | 0.71 |
| 279 | 5 | 289.55 | 5.83 | 66.00 | 9.28 | 0.70 |
| 313 | 5 | 313.00 | 5.00 | 0.00 | 0.00 | 0.00 |
| 395 | 6 | 403.80 | 6.79 | 64.00 | 7.83 | 0.68 |
| 312 | 5 | 321.20 | 5.84 | 65.20 | 8.10 | 0.71 |
| 325 | 5 | 339.10 | 5.84 | 66.60 | 13.35 | 0.69 |
| 358 | 12 | 361.94 | 12.22 | 20.00 | 8.49 | 0.45 |
| 317 | 11 | 320.08 | 11.17 | 15.70 | 7.64 | 0.40 |
| 275 | 4 | 278.58 | 4.44 | 43.40 | 4.27 | 0.50 |
| 395 | 5 | 395.06 | 5.01 | 0.60 | 0.94 | 0.08 |
| 478 | 8 | 479.32 | 8.03 | 3.00 | 8.13 | 0.17 |
| 544 | 7 | 544.22 | 7.01 | 1.00 | 2.52 | 0.10 |
| 100 | 5 | 100.03 | 5.00 | 0.50 | 0.42 | 0.07 |
| 58 | 4 | 58.02 | 4.01 | 0.90 | 0.19 | 0.09 |
| 129 | 6 | 129.03 | 6.01 | 0.70 | 0.38 | 0.08 |
| 91 | 6 | 91.00 | 6.00 | 0.10 | 0.06 | 0.03 |
| 164 | 5 | 164.03 | 5.00 | 0.30 | 0.55 | 0.05 |
| 55 | 3 | 57.21 | 3.69 | 58.10 | 2.66 | 0.65 |
| 123 | 5 | 123.02 | 5.00 | 0.50 | 0.28 | 0.07 |
| 158 | 7 | 158.02 | 7.01 | 1.10 | 0.21 | 0.10 |
| 202 | 9 | 202.09 | 9.04 | 3.90 | 0.46 | 0.19 |
| 235 | 11 | 235.83 | 11.39 | 33.70 | 1.35 | 0.60 |

**Table 6.** Minimisation of $\overline{H} + k.\sigma(H)$

| $h(S)$ | $t(S)$ | $\overline{H}(S,n)$ | $\overline{T}(S,n)$ | $p(S,n)$ | $\sigma_H(S,n)$ | $\sigma_T(S,n)$ |
|---|---|---|---|---|---|---|
| 337 | 6 | 337.00 | 6.00 | 0.00 | 0.00 | 0.00 |
| 366 | 7 | 366.00 | 7.00 | 0.00 | 0.00 | 0.00 |
| 296 | 6 | 296.00 | 6.00 | 0.00 | 0.00 | 0.00 |
| 313 | 5 | 313.00 | 5.00 | 0.00 | 0.00 | 0.00 |
| 409 | 7 | 409.00 | 7.00 | 0.00 | 0.00 | 0.00 |
| 324 | 6 | 324.00 | 6.00 | 0.00 | 0.00 | 0.00 |
| 351 | 6 | 351.00 | 6.00 | 0.00 | 0.00 | 0.00 |
| 388 | 13 | 388.10 | 13.00 | 0.40 | 1.59 | 0.06 |
| 335 | 12 | 335.07 | 12.01 | 0.50 | 1.23 | 0.07 |
| 283 | 5 | 283.00 | 5.00 | 0.00 | 0.00 | 0.00 |
| 403 | 6 | 403.02 | 6.00 | 0.10 | 0.70 | 0.03 |
| 534 | 8 | 534.00 | 8.04 | 4.10 | 0.00 | 0.20 |
| 552 | 8 | 552.00 | 8.00 | 0.00 | 0.00 | 0.00 |
| 102 | 5 | 102.04 | 5.04 | 4.00 | 0.54 | 0.20 |
| 58 | 4 | 58.00 | 4.00 | 0.00 | 0.00 | 0.00 |
| 129 | 6 | 129.00 | 6.00 | 0.00 | 0.00 | 0.00 |
| 91 | 5 | 91.00 | 5.00 | 0.00 | 0.00 | 0.00 |
| 164 | 5 | 164.01 | 5.00 | 0.10 | 0.25 | 0.03 |
| 63 | 4 | 63.00 | 4.00 | 0.00 | 0.00 | 0.00 |
| 123 | 5 | 123.00 | 5.00 | 0.00 | 0.00 | 0.00 |
| 158 | 7 | 158.01 | 7.00 | 0.20 | 0.34 | 0.04 |
| 202 | 9 | 202.00 | 9.00 | 0.20 | 0.09 | 0.04 |
| 238 | 12 | 238.01 | 12.01 | 0.90 | 0.15 | 0.09 |

## APPENDIX 2: BELENGUER AND BENAVENT'S INSTANCES

**Table 7.** Slack approach

| | $h(S)$ | $t(S)$ | $\overline{H}(S,n)$ | $\overline{T}(S,n)$ | $p(S,n)$ | $\sigma_H(S,n)$ | $\sigma_T(S,n)$ |
|---|---|---|---|---|---|---|---|
| val1a | 173 | 2 | 173.0 | 2.0 | 0.0 | 0.0 | 0.00 |
| val1b | 179 | 4 | 179.0 | 4.0 | 0.2 | 0.6 | 0.04 |
| val1c | 261 | 10 | 261.2 | 10.0 | 1.4 | 2.0 | 0.12 |
| val2a | 227 | 2 | 227.0 | 2.0 | 0.0 | 0.0 | 0.00 |
| val2b | 262 | 3 | 262.0 | 3.0 | 0.0 | 0.0 | 0.00 |
| val2c | 499 | 9 | 501.8 | 9.1 | 8.8 | 10.5 | 0.30 |
| val3a | 81 | 2 | 81.0 | 2.0 | 0.0 | 0.0 | 0.00 |
| val3b | 89 | 4 | 89.0 | 4.0 | 0.2 | 0.2 | 0.04 |
| val3c | 146 | 8 | 147.0 | 8.1 | 5.1 | 4.5 | 0.23 |
| val4a | 406 | 4 | 406.0 | 4.0 | 0.0 | 0.0 | 0.00 |
| val4b | 418 | 5 | 418.0 | 5.0 | 0.0 | 0.0 | 0.00 |
| val4c | 448 | 6 | 448.0 | 6.0 | 0.0 | 0.0 | 0.00 |
| val4d | 565 | 10 | 565.5 | 10.0 | 1.0 | 4.9 | 0.10 |
| val5a | 441 | 4 | 441.0 | 4.0 | 0.0 | 0.0 | 0.00 |
| val5b | 466 | 5 | 466.0 | 5.0 | 0.0 | 0.0 | 0.00 |
| val5c | 497 | 6 | 497.0 | 6.0 | 0.0 | 0.0 | 0.00 |
| val5d | 617 | 10 | 617.2 | 10.0 | 0.8 | 2.7 | 0.09 |
| val6a | 223 | 3 | 223.0 | 3.0 | 0.0 | 0.0 | 0.00 |
| val6b | 245 | 5 | 245.0 | 5.0 | 0.1 | 0.3 | 0.03 |
| val6c | 337 | 11 | 337.4 | 11.0 | 2.4 | 2.8 | 0.15 |
| val7a | 279 | 4 | 279.0 | 4.0 | 0.0 | 0.0 | 0.00 |
| val7b | 283 | 5 | 283.0 | 5.0 | 0.0 | 0.0 | 0.00 |
| val7c | 352 | 10 | 352.1 | 10.0 | 0.7 | 1.3 | 0.08 |
| val8a | 395 | 4 | 395.0 | 4.0 | 0.0 | 0.0 | 0.00 |
| val8b | 409 | 5 | 409.0 | 5.0 | 0.0 | 0.0 | 0.00 |
| val8c | 565 | 10 | 565.2 | 10.0 | 0.7 | 2.4 | 0.08 |
| val9a | 326 | 4 | 326.0 | 4.0 | 0.0 | 0.0 | 0.00 |
| val9b | 332 | 5 | 332.0 | 5.0 | 0.0 | 0.0 | 0.00 |
| val9c | 340 | 6 | 340.0 | 6.0 | 0.0 | 0.0 | 0.00 |
| val9d | 409 | 11 | 409.1 | 11.0 | 0.7 | 1.4 | 0.08 |
| val10a | 436 | 4 | 436.0 | 4.0 | 0.0 | 0.0 | 0.00 |
| val10b | 446 | 5 | 446.0 | 5.0 | 0.0 | 0.0 | 0.00 |
| val10c | 459 | 6 | 459.0 | 6.0 | 0.0 | 0.0 | 0.00 |
| val10d | 555 | 11 | 555.0 | 11.0 | 0.1 | 1.0 | 0.03 |

**Table 8.** Minimisation of $\overline{H}$

| $h(S)$ | $t(S)$ | $\overline{H}(S,n)$ | $\overline{T}(S,n)$ | $p(S,n)$ | $\sigma_H(S,n)$ | $\sigma_T(S,n)$ |
|---|---|---|---|---|---|---|
| 173 | 2 | 173.00 | 2.00 | 0.00 | 0.00 | 0.00 |
| 179 | 4 | 179.03 | 4.00 | 0.40 | 0.46 | 0.06 |
| 248 | 9 | 252.22 | 9.28 | 25.10 | 7.86 | 0.51 |
| 227 | 2 | 227.00 | 2.00 | 0.00 | 0.00 | 0.00 |
| 260 | 3 | 260.14 | 3.00 | 0.30 | 2.52 | 0.05 |
| 491 | 9 | 494.71 | 9.09 | 8.30 | 12.91 | 0.29 |
| 81 | 2 | 81.00 | 2.00 | 0.00 | 0.00 | 0.00 |
| 87 | 3 | 87.03 | 3.00 | 0.50 | 0.42 | 0.07 |
| 146 | 8 | 147.02 | 8.05 | 5.20 | 4.67 | 0.22 |
| 400 | 3 | 400.04 | 3.00 | 0.40 | 0.68 | 0.06 |
| 412 | 4 | 412.03 | 4.00 | 0.50 | 0.62 | 0.07 |
| 444 | 5 | 452.44 | 5.82 | 63.80 | 9.56 | 0.75 |
| 549 | 9 | 567.88 | 9.89 | 68.10 | 25.00 | 0.75 |
| 423 | 3 | 423.04 | 3.00 | 0.30 | 0.80 | 0.05 |
| 446 | 4 | 446.09 | 4.01 | 1.20 | 0.99 | 0.11 |
| 474 | 5 | 476.35 | 5.28 | 25.80 | 6.07 | 0.49 |
| 605 | 9 | 606.76 | 9.10 | 9.20 | 6.59 | 0.31 |
| 223 | 3 | 223.02 | 3.00 | 0.20 | 0.54 | 0.04 |
| 235 | 4 | 236.75 | 4.33 | 32.70 | 3.58 | 0.49 |
| 331 | 10 | 333.84 | 10.44 | 38.00 | 4.86 | 0.61 |
| 279 | 4 | 279.00 | 4.00 | 0.00 | 0.00 | 0.00 |
| 283 | 5 | 283.02 | 5.00 | 0.40 | 0.38 | 0.06 |
| 342 | 10 | 344.9 | 10.22 | 20.30 | 7.39 | 0.47 |
| 386 | 3 | 386.18 | 3.02 | 1.80 | 1.36 | 0.13 |
| 395 | 4 | 395.52 | 4.05 | 5.40 | 2.41 | 0.23 |
| 558 | 10 | 559.14 | 10.06 | 6.00 | 5.67 | 0.25 |
| 323 | 3 | 323.00 | 3.00 | 0.00 | 0.00 | 0.00 |
| 326 | 4 | 326.04 | 4.01 | 0.60 | 0.57 | 0.08 |
| 332 | 5 | 332.18 | 5.02 | 2.30 | 1.26 | 0.15 |
| 409 | 11 | 409.02 | 11.00 | 0.10 | 0.51 | 0.03 |
| 428 | 3 | 428.06 | 3.01 | 1.40 | 0.47 | 0.12 |
| 436 | 4 | 436.00 | 4.00 | 0.00 | 0.00 | 0.00 |
| 448 | 5 | 448.23 | 5.04 | 3.60 | 1.37 | 0.19 |
| 541 | 10 | 544.79 | 10.36 | 30.90 | 7.03 | 0.58 |

**Table 9.** Minimisation of $\overline{H} + k.\sigma(H)$

| $h(S)$ | $t(S)$ | $\overline{H}(S,n)$ | $\overline{T}(S,n)$ | $p(S,n)$ | $\sigma_H(S,n)$ | $\sigma_T(S,n)$ |
|---|---|---|---|---|---|---|
| 173 | 2 | 173.00 | 2.00 | 0.00 | 0.00 | 0.00 |
| 179 | 4 | 179.00 | 4.00 | 0.00 | 0.00 | 0.00 |
| 268 | 10 | 268.03 | 10.00 | 0.20 | 0.60 | 0.04 |
| 227 | 2 | 227.00 | 2.00 | 0.00 | 0.00 | 0.00 |
| 260 | 3 | 260.00 | 3.00 | 0.00 | 0.00 | 0.00 |
| 568 | 10 | 568.00 | 10.00 | 0.00 | 0.00 | 0.00 |
| 81 | 2 | 81.00 | 2.00 | 0.00 | 0.00 | 0.00 |
| 87 | 3 | 87.02 | 3.00 | 0.30 | 0.36 | 0.05 |
| 162 | 9 | 162.09 | 9.00 | 0.40 | 1.39 | 0.06 |
| 400 | 3 | 400.00 | 3.00 | 0.00 | 0.00 | 0.00 |
| 422 | 4 | 422.03 | 4.00 | 0.30 | 0.63 | 0.05 |
| 454 | 6 | 454.07 | 6.00 | 0.30 | 1.40 | 0.05 |
| 574 | 10 | 576.41 | 10.21 | 20.30 | 8.04 | 0.42 |
| 428 | 3 | 428.00 | 3.00 | 0.00 | 0.00 | 0.00 |
| 450 | 4 | 450.24 | 4.21 | 21.50 | 1.84 | 0.41 |
| 495 | 6 | 495.00 | 6.00 | 0.00 | 0.00 | 0.00 |
| 631 | 10 | 631.11 | 10.00 | 0.30 | 2.07 | 0.05 |
| 223 | 3 | 223.00 | 3.00 | 0.00 | 0.00 | 0.00 |
| 245 | 5 | 245.00 | 5.00 | 0.00 | 0.00 | 0.00 |
| 347 | 12 | 347.14 | 12.00 | 0.40 | 2.26 | 0.06 |
| 279 | 4 | 279.00 | 4.00 | 0.00 | 0.00 | 0.00 |
| 283 | 5 | 283.00 | 5.00 | 0.00 | 0.00 | 0.00 |
| 352 | 10 | 352.03 | 10.00 | 0.30 | 0.55 | 0.05 |
| 392 | 3 | 392.19 | 3.31 | 31.30 | 1.22 | 0.46 |
| 409 | 5 | 409.08 | 5.00 | 0.30 | 1.46 | 0.05 |
| 583 | 11 | 583.04 | 11.00 | 0.10 | 1.33 | 0.03 |
| 323 | 3 | 323.00 | 3.00 | 0.00 | 0.00 | 0.00 |
| 332 | 5 | 332.00 | 5.00 | 0.00 | 0.00 | 0.00 |
| 340 | 6 | 340.00 | 6.00 | 0.00 | 0.00 | 0.00 |
| 421 | 11 | 421.06 | 11.00 | 0.40 | 0.92 | 0.06 |
| 428 | 3 | 428.02 | 3.00 | 0.20 | 0.46 | 0.04 |
| 436 | 4 | 436.00 | 4.00 | 0.00 | 0.00 | 0.00 |
| 459 | 6 | 459.00 | 6.00 | 0.00 | 0.00 | 0.00 |
| 562 | 11 | 562.00 | 11.00 | 0.00 | 0.00 | 0.00 |

STOCHASTIC CAPACITATED ARC ROUTING PROBLEM*## APPENDIX 3: EGLESE'S INSTANCES

**Table 10.** S*lack* approach

| | $h(S)$ | $t(S)$ | $\overline{H}(S,n)$ | $\overline{T}(S,n)$ | $p(S,n)$ | $s_H(S,n)$ | $s_T(S,n)$ |
|---|---|---|---|---|---|---|---|
| egl-e1-A | 3825 | 6 | 3833.31 | 6.01 | 1.50 | 68.36 | 0.12 |
| egl-e1-B | 4805 | 8 | 4822.79 | 8.04 | 4.40 | 85.25 | 0.21 |
| egl-e1-C | 6161 | 11 | 6219.58 | 11.11 | 10.70 | 186.29 | 0.32 |
| egl-e2-A | 5418 | 8 | 5430.94 | 8.03 | 2.60 | 83.83 | 0.16 |
| egl-e2-B | 6791 | 11 | 6809.15 | 11.04 | 4.30 | 90.20 | 0.20 |
| egl-e2-C | 9369 | 16 | 9467.07 | 16.2 | 18.70 | 223.90 | 0.44 |
| egl-e3-A | 6355 | 9 | 6361.31 | 9.01 | 1.30 | 58.99 | 0.13 |
| egl-e3-B | 8590 | 13 | 8653.53 | 13.14 | 13.70 | 167.20 | 0.37 |
| egl-e3-C | 11412 | 19 | 11488.62 | 19.17 | 16.00 | 187.27 | 0.40 |
| egl-e4-A | 6989 | 10 | 6998.36 | 10.02 | 1.80 | 70.61 | 0.13 |
| egl-e4-B | 9791 | 16 | 9861.22 | 16.15 | 14.10 | 191.62 | 0.37 |
| egl-e4-C | 12864 | 22 | 13037.68 | 22.35 | 30.00 | 309.80 | 0.60 |
| egl-s1-A | 5502 | 8 | 5511.16 | 8.02 | 2.40 | 72.37 | 0.15 |
| egl-s1-B | 6913 | 11 | 6934.01 | 11.05 | 4.60 | 102.71 | 0.21 |
| egl-s1-C | 9318 | 16 | 9350.22 | 16.06 | 5.30 | 146.75 | 0.24 |
| egl-s2-A | 10927 | 16 | 10934.66 | 16.01 | 1.40 | 65.72 | 0.12 |
| egl-s2-B | 14660 | 23 | 14738.68 | 23.12 | 11.30 | 230.59 | 0.33 |
| egl-s2-C | 18238 | 30 | 18513.7 | 30.51 | 41.90 | 406.62 | 0.68 |
| egl-s3-A | 11149 | 16 | 11162.94 | 16.03 | 2.70 | 90.42 | 0.16 |
| egl-s3-B | 15140 | 24 | 15211.26 | 24.19 | 16.90 | 188.45 | 0.43 |
| egl-s3-C | 19076 | 32 | 19310.07 | 32.54 | 42.50 | 369.37 | 0.71 |
| egl-s4-A | 13411 | 21 | 13437.76 | 21.05 | 5.40 | 124.01 | 0.23 |
| egl-s4-B | 18117 | 30 | 18203.24 | 30.20 | 17.70 | 212.14 | 0.45 |
| egl-s4-C | 22827 | 40 | 23077.83 | 40.51 | 40.30 | 392.21 | 0.71 |

**Table 11.** Minimisation of $\overline{H}$

| $h(S)$ | $t(S)$ | $\overline{H}(S,n)$ | $\overline{T}(S,n)$ | $p(S,n)$ | $s_H(S,n)$ | $s_T(S,n)$ |
|---|---|---|---|---|---|---|
| 3568 | 5 | 3699.39 | 5.91 | 66.10 | 183.20 | 0.81 |
| 4659 | 8 | 4752.43 | 8.39 | 34.70 | 149.42 | 0.57 |
| 6043 | 11 | 6136.79 | 11.61 | 53.20 | 184.20 | 0.63 |
| 5152 | 7 | 5354.12 | 8.16 | 75.20 | 220.04 | 0.91 |
| 6541 | 11 | 6719.82 | 11.83 | 64.10 | 268.22 | 0.75 |
| 8794 | 15 | 9134.37 | 15.70 | 53.00 | 406.31 | 0.78 |
| 6383 | 9 | 6383.19 | 9.00 | 0.30 | 3.50 | 0.05 |
| 8357 | 13 | 8726.64 | 13.59 | 53.20 | 369.72 | 0.61 |
| 11418 | 19 | 11481.11 | 19.14 | 13.00 | 172.92 | 0.37 |
| 6866 | 10 | 6995.42 | 10.66 | 54.70 | 192.32 | 0.67 |
| 9731 | 15 | 9847.92 | 15.78 | 59.00 | 187.42 | 0.77 |
| 12717 | 21 | 13145.49 | 22.61 | 86.20 | 399.69 | 1.06 |
| 5160 | 7 | 5337.34 | 7.65 | 50.10 | 224.70 | 0.74 |
| 6726 | 10 | 6912.86 | 10.99 | 65.30 | 279.10 | 0.92 |
| 9044 | 15 | 9403.60 | 15.77 | 59.20 | 441.27 | 0.77 |
| 10506 | 15 | 10646.39 | 15.34 | 29.80 | 243.03 | 0.57 |
| 14064 | 22 | 14375.76 | 22.64 | 49.60 | 433.37 | 0.76 |
| 17736 | 29 | 18468.76 | 30.89 | 88.40 | 622.76 | 1.23 |
| 10876 | 15 | 11105.16 | 16.43 | 81.80 | 258.11 | 1.00 |
| 14858 | 23 | 15449.24 | 24.94 | 90.60 | 468.25 | 1.18 |
| 18589 | 31 | 19349.71 | 33.39 | 94.70 | 528.54 | 1.25 |
| 13418 | 20 | 13642.57 | 20.62 | 47.70 | 323.59 | 0.75 |
| 17504 | 29 | 18112.71 | 30.50 | 84.60 | 611.50 | 1.05 |
| 22200 | 38 | 23394.89 | 40.88 | 95.90 | 788.90 | 1.58 |

**Table 12.** Minimisation of $\overline{H}+k.\sigma(H)$

| $h(S)$ | $t(S)$ | $\overline{H}(S,n)$ | $\overline{T}(S,n)$ | $p(S,n)$ | $s_H(S,n)$ | $s_T(S,n)$ |
|---|---|---|---|---|---|---|
| 4036 | 6 | 4036.00 | 6.00 | 0.00 | 0.00 | 0.00 |
| 4931 | 8 | 4932.70 | 8.01 | 0.70 | 26.61 | 0.08 |
| 7029 | 12 | 7030.88 | 12.01 | 0.50 | 28.68 | 0.07 |
| 5628 | 8 | 5629.08 | 8.00 | 0.30 | 19.74 | 0.05 |
| 7719 | 12 | 7720.10 | 12.00 | 0.20 | 24.48 | 0.04 |
| 10534 | 18 | 10534.56 | 18.00 | 0.10 | 17.76 | 0.03 |
| 6625 | 9 | 6627.11 | 9.01 | 0.50 | 30.17 | 0.07 |
| 10262 | 15 | 10264.30 | 15.01 | 0.50 | 32.71 | 0.07 |
| 12060 | 20 | 12066.39 | 20.01 | 1.40 | 54.51 | 0.12 |
| 7754 | 11 | 7756.54 | 11.00 | 0.40 | 41.58 | 0.06 |
| 11164 | 17 | 11168.41 | 17.02 | 1.50 | 44.28 | 0.12 |
| 15426 | 25 | 15427.97 | 25.01 | 1.00 | 28.21 | 0.10 |
| 6320 | 9 | 6320.57 | 9.00 | 0.10 | 18.08 | 0.03 |
| 8109 | 11 | 8113.54 | 11.37 | 37.10 | 34.37 | 0.49 |
| 10957 | 17 | 10957.61 | 17.00 | 0.50 | 9.49 | 0.07 |
| 11289 | 15 | 11390.78 | 16.32 | 83.90 | 146.21 | 0.84 |
| 17376 | 25 | 17403.18 | 25.53 | 51.00 | 41.65 | 0.54 |
| 21412 | 32 | 21507.33 | 32.55 | 49.70 | 160.34 | 0.60 |
| 12107 | 17 | 12118.60 | 17.06 | 6.10 | 69.95 | 0.26 |
| 18708 | 27 | 18717.61 | 27.02 | 2.10 | 74.32 | 0.14 |
| 22803 | 36 | 22847.58 | 36.13 | 12.90 | 139.37 | 0.36 |
| 14355 | 21 | 14433.81 | 21.98 | 71.60 | 117.80 | 0.78 |
| 21421 | 33 | 21503.67 | 33.38 | 36.00 | 189.53 | 0.53 |
| 27168 | 43 | 27226.95 | 43.51 | 48.00 | 124.99 | 0.55 |

## APPENDIX 4: COMPUTATIONAL TIME

**Table 13.** Computational time in secondes for DeArmon's instances

|  | Minimisation of $\overline{H}$ | | Minimisation of $\overline{H}+k.\sigma$ | |
|---|---|---|---|---|
|  | Total Time | Time to Best | Total Time | Time to Best |
| gdb1  | 61.08  | 8.84  | 49.84  | 0.56 |
| gdb2  | 63.13  | 4.59  | 75.33  | 6.73 |
| gdb3  | 55.56  | 3.44  | 57.76  | 1.2 |
| gdb4  | 44.23  | 40.81 | 50.75  | 4.05 |
| gdb5  | 62.39  | 3.61  | 69.77  | 2.58 |
| gdb6  | 54.92  | 7.08  | 59.39  | 0.2 |
| gdb7  | 53.08  | 0.83  | 54.08  | 11.7 |
| gdb8  | 123    | 90.26 | 130.61 | 74.8 |
| gdb9  | 137.25 | 48.37 | 159.64 | 159.36 |
| gdb10 | 54.26  | 6.3   | 54.67  | 2.03 |
| gdb11 | 116.55 | 9.36  | 114.75 | 4.66 |
| gdb12 | 56.75  | 21.81 | 65.39  | 0.47 |
| gdb13 | 60.92  | 41.02 | 66.42  | 14.59 |
| gdb14 | 49.72  | 3.97  | 50.69  | 29.03 |
| gdb15 | 68.02  | 0.12  | 58.56  | 0.52 |
| gdb16 | 60.08  | 1.64  | 73.7   | 6.81 |
| gdb17 | 66.97  | 0.12  | 72.3   | 0.33 |
| gdb18 | 81.64  | 2.23  | 86.84  | 0.28 |
| gdb19 | 37.88  | 0.09  | 32.51  | 0.02 |
| gdb20 | 48.12  | 1.53  | 57.06  | 1.44 |
| gdb21 | 71.55  | 1.84  | 96.33  | 14.62 |
| gdb22 | 108.44 | 25.22 | 115    | 86.48 |
| gdb23 | 129.81 | 92.05 | 153.78 | 152.41 |
| Avg.  | 72.41  | 18.05 | 78.49  | 24.99 |

**Table 14.** Computational time in secondes for Belenguer and Benavent's instances

|        | Minimisation of $\overline{H}$ | | Minimisation of $\overline{H}+k.\sigma$ | |
|--------|------------|--------------|------------|--------------|
|        | Total Time | Time to Best | Total Time | Time to Best |
| val1a  | 0.05    | 0.03   | 0.03   | 0.03   |
| val1b  | 100.36  | 1.2    | 110.92 | 0.25   |
| val1c  | 95.7    | 30.59  | 130.25 | 10.23  |
| val2a  | 92.25   | 0.33   | 99.39  | 0.2    |
| val2b  | 84.75   | 0.7    | 92.39  | 26.03  |
| val2c  | 87.78   | 9.92   | 118.51 | 17.98  |
| val3a  | 99.8    | 0.19   | 103.27 | 0.2    |
| val3b  | 93.16   | 3.48   | 101.89 | 12.75  |
| val3c  | 83.52   | 51.56  | 91.31  | 1.62   |
| val4a  | 263.59  | 9.8    | 264.53 | 9.2    |
| val4b  | 234.41  | 7.66   | 271.33 | 13.34  |
| val4c  | 227.97  | 81.91  | 257.97 | 153.83 |
| val4d  | 221.89  | 167.66 | 243.06 | 27.31  |
| val5a  | 234.89  | 14     | 242.28 | 15.58  |
| val5b  | 246.17  | 50.88  | 227.87 | 114.08 |
| val5c  | 214.31  | 91.98  | 207.59 | 61.36  |
| val5d  | 215.73  | 84.55  | 217.92 | 104.03 |
| val6a  | 157.83  | 7.86   | 162.7  | 0.62   |
| val6b  | 142.27  | 14.58  | 142.28 | 1.94   |
| val6c  | 143.64  | 116.69 | 155.95 | 7.13   |
| val7a  | 228.42  | 2.33   | 240.16 | 56.05  |
| val7b  | 215.53  | 2.56   | 226.27 | 2.19   |
| val7c  | 177.91  | 177.41 | 197.42 | 98.14  |
| val8a  | 228.12  | 38.2   | 222.98 | 205.05 |
| val8b  | 210.98  | 22.61  | 219.95 | 16.95  |
| val8c  | 165.7   | 62.97  | 208.3  | 30.63  |
| val9a  | 540.08  | 80.23  | 520.55 | 265.56 |
| val9b  | 444.55  | 389.77 | 436.22 | 5.34   |
| val9c  | 464.69  | 64.42  | 398.73 | 24.13  |
| val9d  | 348.77  | 149.31 | 380.39 | 176    |
| val10a | 583.06  | 58.03  | 582.36 | 149.66 |
| val10b | 444.91  | 267.69 | 479.23 | 34.56  |
| val10c | 411.61  | 64.33  | 476.05 | 23.22  |
| val10d | 389.55  | 195.91 | 398.36 | 361.52 |
| Avg.   | 289.50  | 95.91  | 297.59 | 84.13  |

**Table 15.** Computational time in secondes for Eglese's instances

|          | Minimisation of $\overline{H}$ | | Minimisation of $\overline{H} + k.\sigma$ | |
|----------|-----------:|-----------:|-----------:|-----------:|
|          | Total Time | Time to Best | Total Time | Time to Best |
| egl-e1-A | 220.37  | 106.01  | 168.34  | 165.27 |
| egl-e1-B | 168.36  | 147.83  | 142.28  | 8.48 |
| egl-e1-C | 164.83  | 139.66  | 180.03  | 128.97 |
| egl-e2-A | 260.84  | 96.7    | 233.41  | 212.97 |
| egl-e2-B | 290.92  | 276.83  | 370.75  | 181.91 |
| egl-e2-C | 284.81  | 247.75  | 264.8   | 116.94 |
| egl-e3-A | 338.67  | 55.09   | 357.27  | 257.77 |
| egl-e3-B | 441.53  | 441.34  | 587.02  | 65.25 |
| egl-e3-C | 377.8   | 351.45  | 570.34  | 187.33 |
| egl-e4-A | 537.11  | 536.48  | 449.23  | 121.34 |
| egl-e4-B | 503.23  | 500.17  | 466.8   | 319.55 |
| egl-e4-C | 408.47  | 322.74  | 563.05  | 27.89 |
| egl-s1-A | 347.76  | 305.13  | 328.73  | 225.05 |
| egl-s1-B | 301.2   | 189.58  | 386.83  | 98.58 |
| egl-s1-C | 311.03  | 127.02  | 296.06  | 211.92 |
| egl-s2-A | 1133.45 | 900.28  | 1189.11 | 781.69 |
| egl-s2-B | 962.76  | 828.7   | 987.31  | 908.44 |
| egl-s2-C | 1425.81 | 411.7   | 934.42  | 724.89 |
| egl-s3-A | 1176.33 | 680.09  | 1018.59 | 820.91 |
| egl-s3-B | 906.84  | 842.89  | 1220.84 | 1159.17 |
| egl-s3-C | 1111.72 | 501.91  | 1060.11 | 1047.59 |
| egl-s4-A | 1462.44 | 1234.86 | 1398.59 | 585.16 |
| egl-s4-B | 1489.88 | 1289.2  | 3467.5  | 2727.91 |
| egl-s4-C | 1685.25 | 1537.05 | 1720.36 | 1331.92 |
| Avg.     | 699.61  | 520.19  | 791.02  | 532.68 |